\documentclass{article}

\usepackage{arxiv}

\usepackage[utf8]{inputenc} 
\usepackage[T1]{fontenc}    
\usepackage{hyperref}       
\usepackage{url}            
\usepackage{booktabs}       
\usepackage{amsfonts}       
\usepackage{nicefrac}       
\usepackage{microtype}      
\usepackage{lipsum}		
\usepackage{graphicx}
\usepackage[numbers]{natbib}
\usepackage{doi}
\usepackage{amsmath,amssymb,amsfonts}
\usepackage{algorithmic}
\usepackage{graphicx}
\usepackage{textcomp}
\usepackage{multirow}
\usepackage{multicol}
\usepackage{booktabs}

\hypersetup{hidelinks,
	colorlinks=true,
	allcolors=black,
	pdfstartview=Fit,
	breaklinks=true}

\title{Semantic-Aware Subgraph State Space Model for WSI Classification in Histopathology}

\date{31 Lúnasa, 2026}	

\author{ \href{https://orcid.org/0009-0007-5035-1051}{\includegraphics[scale=0.06]{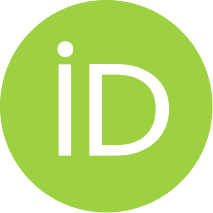}\hspace{1mm} Feixing ~Chen} \\
	  National College for Excellent Engineers\\
	Beihang University\\
	Beijing, P.R.C. 100191 \\
	\texttt{feixing\_chen@buaa.edu.cn} \\
	\And
    \includegraphics[scale=0.06]{orcid.pdf}\hspace{1mm}Hao~Lu \\
	School of Biological Science and Medical Engineering\\
	Beihang University\\
	Beijing, P.R.C. 100191 \\
	\texttt{lu2002@buaa.edu.cn} \\
    \And
	\href{https://orcid.org/0000-0001-5752-329X}{\includegraphics[scale=0.06]{orcid.pdf}\hspace{1mm}Lin~Luo}\thanks{Corresponding Author} \\
	College of Engineering\\
	Peking University\\
	Beijing, P.R.C. 100871 \\
	\texttt{luol@pku.edu.cn} \\
    \And
	\href{https://orcid.org/0000-0002-2636-7594}{\includegraphics[scale=0.06]{orcid.pdf}\hspace{1mm}Yan~Xu}\textsuperscript{*} \\
	School of Biological Science and Medical Engineering\\
	Beihang University\\
	Beijing, P.R.C. 100191 \\
	\texttt{xuyan04@gmail.com} \\
}

\renewcommand{\shorttitle}{\textit{arXiv} Template}

\hypersetup{
pdftitle={A template for the arxiv style},
pdfsubject={q-bio.NC, q-bio.QM},
pdfauthor={David S.~Hippocampus, Elias D.~Striatum},
pdfkeywords={First keyword, Second keyword, More},
}

\begin{document}
\maketitle

\begin{abstract}
Histopathological subtyping relies on the recognition of characteristic histological patterns. 
These patterns may be expressed by individual tissue structures or by the spatial distribution and co-occurrence of multiple structures, and they often span irregularly shaped tissue regions, termed semantic units in this work.
However, conventional patch-based representations may fragment such units and fail to explicitly preserve their internal spatial organization, while efficiently modeling relationships among numerous spatially separated units remains challenging.
To address these limitations, we propose the Semantic-Aware Subgraph State Space Model (SASG-SSM), a flexible and efficient framework for whole slide image (WSI) classification. 
Semantic-Aware Subgraphs (SASGs) first approximate irregularly shaped semantic units by adaptively grouping spatially connected patches guided by class-agnostic visual-semantic priors.
By representing patches as graph nodes with adjacency edges, SASGs preserve their internal spatial organization rather than treating them as an unordered set.
A Subgraph State Space Module (SG-SSM) subsequently combines a graph neural network encoder for intra-subgraph topology encoding with a Mamba-based state space encoder for efficient contextualization across large numbers of subgraphs. 
This module integrates local structural information within semantic units with global contextual information arising from their distribution and co-occurrence across the WSI, while efficiently modeling a large number of spatially distributed regions.
Extensive experiments across four WSI subtyping datasets demonstrate consistent advantages over representative state-of-the-art methods. Further evaluations under small-cohort and few-shot settings demonstrate robustness and data efficiency under limited training data.
Code will be released at https://github.com/HLSvois/SASG-SSM.
\end{abstract}


\section{Introduction}
Histopathological subtyping is essential for treatment selection and prognostic assessment~\cite{breen_comprehensive_2025}. 
It relies on characteristic histological patterns, including tumor nests, malignant glands, papillary or tubular formations, keratinization, and spatial arrangements of tumor and surrounding tissues. 
For example, papillary architecture is characteristic of papillary renal cell carcinoma, whereas invasive squamous nests with varying degrees of keratinization are commonly exhibited in esophageal squamous cell carcinoma~\cite{fletcher_diagnostic_2013},~\cite{who_classification_of_tumours_editorial_board_digestive_2019}. 
These patterns often span irregularly shaped tissue regions and can be characterized by both the morphology of individual structures and their spatial distribution and co-occurrence~\cite{chen_spatial_2024}. 
In this work, we use the term \emph{semantic unit} to denote a spatially connected and histologically coherent region that conveys a recognizable histological pattern.

Whole slide images (WSIs) are gigapixel-scale digital images widely used in histopathology~\cite{ghaznavi_digital_2013}. 
Owing to computational constraints, conventional WSI pipelines divide foreground tissue into thousands of fixed square patches~\cite{lu_data-efficient_2021}. 
However, an individual patch provides only a limited field of view and may capture merely a fragment of an irregular semantic unit.
Sub-bag methods group multiple patches to enlarge the receptive field~\cite{zhang_dtfd-mil_2022}, but generally treat their constituents as unordered instances and do not explicitly preserve internal spatial organization. 
Graph-based methods could encode relationships among patches using spatial or feature-space edges~\cite{chen_whole_2021},~\cite{zheng_graph-transformer_2022},~\cite{shi_structure-aware_2023}. Although these graphs preserve patch-level relationships, patches typically remain the fundamental modeling entities rather than integrated multi-patch histological structures.

\begin{figure}[!h]
\centerline{\includegraphics[width=0.6\columnwidth]{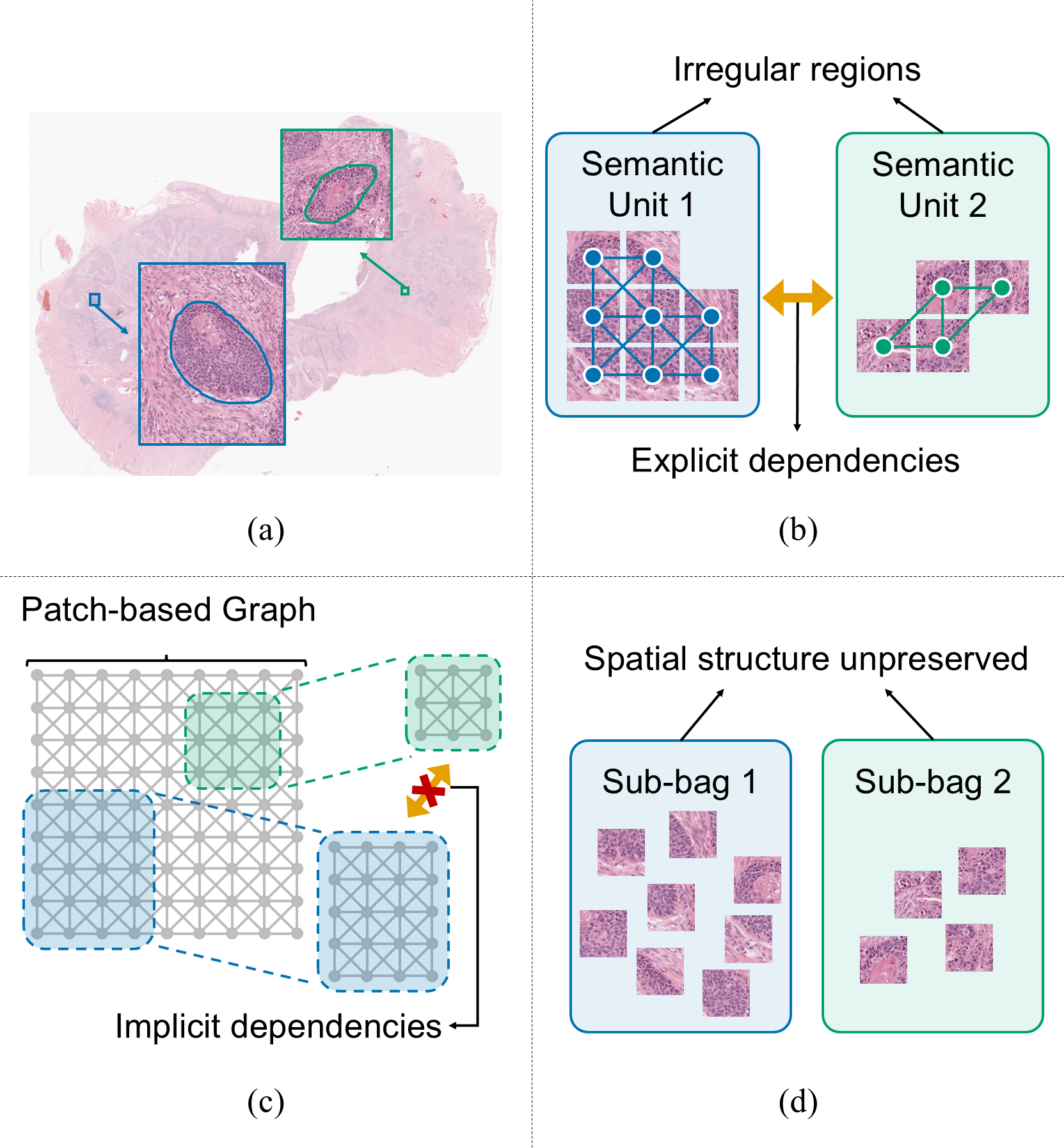}}
\caption{Conceptual comparison of subgraph-, graph-, and sub-bag-based representation strategies. (a) Two representative histological regions selected from a WSI of squamous cell carcinoma.  (b) Subgraph-based strategy represents irregularly shaped semantic units as connected subgraphs and enables contextual dependencies among different units to be modeled explicitly. (c) Graph-based strategy constructs a slide-level graph using patches as nodes. Although patch-level relationships are preserved, multi-patch histological structures are not explicitly represented as integrated modeling units. (d) Sub-bag strategy constructs sub-bags by grouping multiple patches, but their internal spatial organization is not explicitly preserved.}
\label{fig:intro_cmp}
\end{figure}

Two challenges therefore remain. 
First, semantic units often exhibit irregular shapes and variable spatial extents and should be represented while preserving both their extent and internal organization. 
Second, their distribution, co-occurrence, and relationships provide important slide-level context, yet efficiently modeling dependencies among numerous spatially separated units is difficult. 
To address the first challenge, we represent spatially related patches of a semantic unit as a connected subgraph. 
Unlike fixed patches and unordered sub-bags, subgraphs provide irregular spatial support while preserving internal topology, enabling more complete representation of semantic units. Fig.~\ref{fig:intro_cmp} compares sub-bag-, graph-, and subgraph-based strategies.

Beyond individual semantic units, modeling contextual relationships among distant histological regions is important for incorporating slide-level information~\cite{shao_transmil_2021}. 
Transformer-based architectures~\cite{vaswani_attention_2017},~\cite{dosovitskiy_image_2021} have therefore been introduced into WSI analysis, frequently together with down-sampling~\cite{zheng_graph-transformer_2022}, linear approximation~\cite{shao_transmil_2021}, or staged aggregation~\cite{chen_scaling_2022} to alleviate the quadratic computational complexity of self-attention. 
However, they may discard fine-grained regional information or weaken direct information propagation among distant regions. 
Mamba has recently been proposed as a selective state space model (SSM) with input-dependent information propagation and linear computational complexity with respect to sequence length~\cite{gu_mamba_2024}. 
Its ability to efficiently process long sequences makes it suitable for large numbers of histological regions. 
Existing Mamba-based WSI methods mainly focus on patch-level sequence modeling~\cite{yang_mambamil_2024}, ~\cite{huang_unleash_2025}, ~\cite{zhang_2dmamba_2025}, leaving efficient subgraph-level contextualization insufficiently explored.

In this work, we propose a flexible and efficient WSI classification framework termed the Semantic-Aware Subgraph State Space Model (SASG-SSM), consisting of Semantic-Aware Subgraphs (SASGs) and a Subgraph State Space Module (SG-SSM). Inspired by the perspective of histology, SASGs are capable of capturing explicit histologic patterns demonstrated in a certain region, and SG-SSM is specially designed to capture implicit patterns revealed by spatial distribution and co-occurrence of different regions.
SASGs approximate irregularly shaped candidate semantic units by grouping spatially and visually related patches while preserving their internal topology. 
SG-SSM subsequently employs a GNN to encode intra-subgraph structure and a Mamba-based SSM to model long-range contextual dependencies among subgraphs, incorporating both local and global information.
Extensive experiments on four WSI subtyping datasets demonstrate the effectiveness of SASG-SSM, with additional small-cohort and few-shot settings evaluating robustness under limited training data. 
The main contributions are summarized as follows:

\begin{itemize}

\item A flexible and efficient framework for WSI classification, termed the 
\textit{Semantic-Aware Subgraph State Space Model (SASG-SSM)}, is proposed 
to jointly represent histologically meaningful semantic units and model 
contextual dependencies among them.

\item \textit{Semantic-Aware Subgraphs (SASGs)} are introduced to approximate 
candidate semantic units as connected, irregularly shaped multi-patch regions. 
SASGs provide flexible spatial support while preserving intra-unit topology, 
thereby reducing the fragmentation imposed by fixed square patches.

\item The \textit{Subgraph State Space Module (SG-SSM)} incorporates a GNN 
encoder to aggregate intra-subgraph information and a Mamba-based SSM encoder 
to model long-range dependencies among a large number of subgraphs.

\end{itemize}

\section{Related Work}
\label{sec:related_work}

\subsection{WSI Classification}
Weakly supervised multiple instance learning (MIL) is a dominant paradigm for WSI classification, with representative approaches including ABMIL~\cite{ilse_attention-based_2018}, CLAM~\cite{lu_data-efficient_2021}, DSMIL~\cite{li_dual-stream_2021}, and TransMIL~\cite{shao_transmil_2021}. 
These methods typically represent a WSI as a bag of fixed-grid patch features, which may fragment histological structures that do not conform to the grid. 
DTFD-MIL~\cite{zhang_dtfd-mil_2022} further groups patches into pseudo-bags (also known as sub-bags) for double-tier feature distillation under limited supervision. 
However, such sub-bags need not correspond to spatially connected tissue regions and do not explicitly preserve internal spatial organization. 
Instead, our Semantic-Aware Subgraphs (SASGs) organize connected patches as topology-preserving subgraphs with irregular support, providing a structured representation of histological tissues.

\subsection{Graph-Based WSI Analysis}
Graph-based methods model histological structure at different granularities. 
CGC-Net~\cite{zhou_cgc-net_2019} represents nuclei and their interactions, HACT~\cite{pati_hierarchical_2022} jointly models cells and tissue regions, and whole-slide frameworks including PatchGCN~\cite{chen_whole_2021}, Graph-Transformer~\cite{zheng_graph-transformer_2022}, and SGMF~\cite{shi_structure-aware_2023} exploit spatial relationships for slide-level prediction.  
These approaches generally use predefined cells, regions, or clusters as graph entities. 
In contrast, SASG uses a connected irregular multi-patch subgraph itself as the modeling unit while explicitly retaining intra-unit adjacency.

\subsection{State Space Models for WSI Analysis}
State space models (SSMs) provide an efficient alternative for long-sequence modeling~\cite{gu_combining_2021,gu_efficiently_2021}. 
Mamba~\cite{gu_mamba_2024} introduces input-dependent selective state-space parameters with linear complexity, motivating its application to WSI analysis. 
Patch-based methods such as MambaMIL~\cite{yang_mambamil_2024}, PAM~\cite{huang_unleash_2025}, and 2DMamba~\cite{zhang_2dmamba_2025} contextualize patch sequences with different scanning designs.
M3amba~\cite{zheng_m3amba_2025} and GMMamba~\cite{zheng_gmmamba_2025} extend modeling to groups.
And hybrid graph-Mamba frameworks such as GAT-Mamba~\cite{ding_combining_2025} combine graph-based structural modeling with state-space contextualization. 
Nevertheless, these approaches primarily operate on patch-, group-, or graph-node-level representations without fully incorporating local and global information.
In contrast, our Subgraph State Space Module (SG-SSM) encodes topology within each irregular SASG before contextualizing topology-aware subgraph representations with Mamba, establishing a hierarchical local-to-global information flow.

\subsection{Visual Foundation Models}
Visual foundation models (VFMs) are large-scale pretrained models with broad
transferability across downstream vision tasks~\cite{kirillov_segment_2023}.
General-purpose VFMs include CLIP~\cite{radford_learning_2021} and
GLIP~\cite{li_grounded_2022} for transferable visual and vision-language
representations, as well as the Segment Anything family
~\cite{kirillov_segment_2023,ravi_sam_2024,carion_sam_2025} for promptable
segmentation. 
Pathology-specific VFMs, including
UNI~\cite{chen_towards_2024}, CONCH~\cite{lu_visual-language_2024},
Prov-GigaPath~\cite{xu_whole-slide_2024}, and
Virchow~\cite{vorontsov_foundation_2024}, further learn transferable
representations from large-scale histopathology data. 
In this work, we explore
the use of SAM2~\cite{ravi_sam_2024} to generate class-agnostic region masks
as visual-semantic priors, providing coarse boundary cues for constructing
spatially coherent SASGs.

\section{Methodology}
\label{sec:methodology}

\begin{figure*}[!h]
\centering
\includegraphics[width=\textwidth]{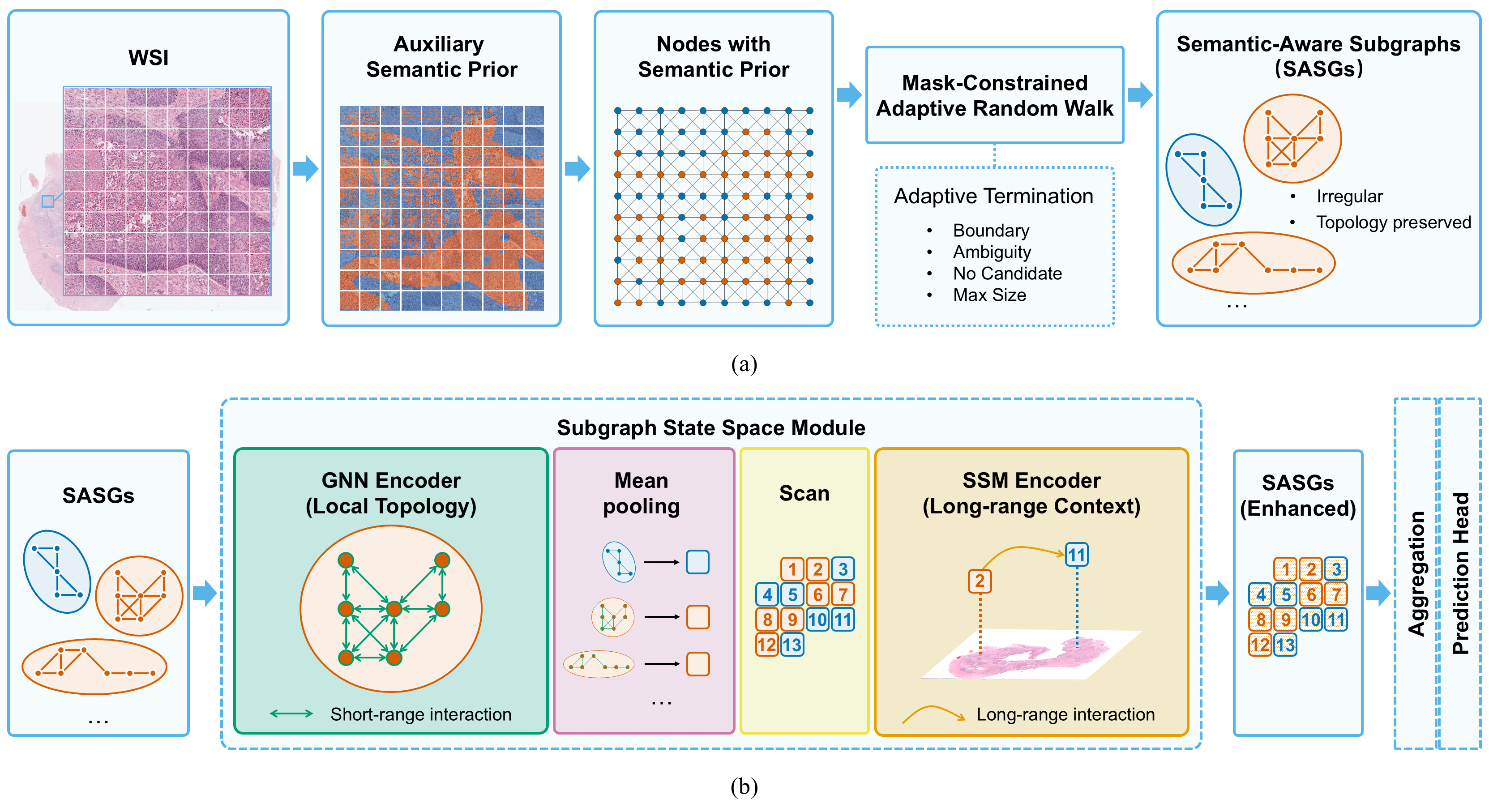}
\caption{Framework of the proposed Semantic-Aware Subgraph State Space Model 
(SASG-SSM). (a) Construction of Semantic-Aware Subgraphs (SASGs). An auxiliary 
visual-semantic prior based on segmentation mask constraints an adaptive random 
walk over patch nodes to construct irregular, topology-preserving subgraphs that 
approximate candidate semantic units. (b) Modeling of Subgraph State Space Module 
(SG-SSM). A GNN encoder first integrates intra-subgraph topological information, 
after which pooled SASG representations are spatially serialized and contextualized 
by a Mamba-based SSM encoder to model long-range contextual dependencies among subgraphs. 
The enhanced representations are subsequently aggregated for slide-level subtype 
classification.}
\label{fig:sasg_ssm}
\end{figure*}

Our Semantic-Aware Subgraph State Space Model (SASG-SSM) is 
illustrated in Fig.~\ref{fig:sasg_ssm}. It comprises two principal components: 
Semantic-Aware Subgraphs (SASGs), which provide irregular, topology-preserving 
representations of semantic units, and a Subgraph State Space Module 
(SG-SSM), which integrates their internal topology and models long-range 
contextual dependencies among them. Together, these components support the 
recognition of histological patterns expressed by tissue structures and their 
combinations across the WSI to facilitate accurate subtype classification.

\subsection{Semantic-Aware Subgraphs}
\label{subsec:sasg}
A semantic unit is a spatially connected and histologically coherent region conveying a recognizable pattern. 
Because such units may have irregular boundaries and variable extents, flexible spatial support can reduce fragmentation and better preserve local structure. 
We therefore organize spatially connected patches into irregularly-shaped, topology-preserving subgraphs, termed Semantic-Aware Subgraphs (SASGs).
Here, ``semantic-aware'' indicates that subgraph has adaptive spatial extent, guided by visual-region cues derived from a class-agnostic semantic prior.

\subsubsection{Auxiliary Semantic Prior}
To guide adaptive subgraph construction
while avoiding unconstrained expansion across visually distinct tissue regions, we first obtain coarse region masks that provide visual boundary cues. Specifically, SAM2~\cite{ravi_sam_2024} is employed to generate a set of class-agnostic masks for each WSI. Given a WSI slide, $R$ masks are derived as the visual-semantic prior:
\begin{equation}
M = \{M_r\}_{r=1}^{R}.
\label{eq:semantic_prior}
\end{equation}

\subsubsection{Irregular Subgraph Sampling}
Fixed square patches impose rigid boundaries that may fragment histological structures with irregular shapes and variable spatial extents.
In contrast, connected subgraphs provide flexible spatial support while
retaining adjacency relationships among their constituent patches. 
Such topology preserves the internal spatial organization of a candidate semantic
unit and enables subsequent GNN to model its local structural arrangement.

Following standard WSI preprocessing~\cite{lu_data-efficient_2021}, the foreground tissue is divided into
$N$ non-overlapping patches and encoded using a pretrained feature extractor $f_{\phi}$:
\begin{equation}
P = \{p_n\}_{n=1}^{N}, \qquad
\mathbf{x}_n = f_{\phi}(p_n) \in \mathbb{R}^{d}, \qquad
c_n \in \mathbb{R}^{2}.
\label{eq:patch_encoding}
\end{equation}
where $p_n$, $\mathbf{x}_n$, and $c_n$ denote the $n$-th patch, its feature 
vector, and its spatial coordinate, respectively, and $d$ represents the 
dimension of feature vector.

A Semantic-Aware Subgraph (SASG) consists of selected $N_s$ patches as nodes 
and builds edges between adjacent patches, thus approximating an irregular 
semantic unit:
\begin{equation}
\begin{aligned}
G_s &= (V_s,\mathbf{X}_s,\mathbf{A}_s), \\
\mathbf{X}_s &\in \mathbb{R}^{N_s \times d}, \qquad
\mathbf{A}_s \in \mathbb{R}^{N_s \times N_s},
\end{aligned}
\label{eq:sasg_definition}
\end{equation}
where $V_s$ denotes the coordinates of selected patches, $\mathbf{X}_s$ denotes 
the feature matrix, and $\mathbf{A}_s$ denotes adjacent matrix of the $s$-th SASG.

To initialize SASG construction, $S=N\times 10\%$ seed patches are randomly selected across the
segmentation masks.
Starting from each seed, a mask-constrained adaptive
random walk progressively selects one unvisited adjacent patch (in its 8-nearest neighbors~\cite{chen_whole_2021}) within the
same mask. 
Because expansion is determined by local connectivity and mask
boundaries rather than a predefined geometric window, the resulting patch set
can assume an irregular shape and variable spatial extent. 
The walk terminates
when any of the following conditions is satisfied:
1) further expansion would leave the seed mask; 
2) the walk has passed low-purity patches for $W_{\mathrm{amb}}=3$ consecutive steps; 
3) no unvisited patch in the adjacent neighborhood; and 
4) the number of selected nodes reaches $W_{\max}=10$.

First condition constrains expansion to the prior region. 
Second condition limits propagation through ambiguous boundaries:
\begin{equation}
q_n =
\max_{r}
\frac{
    \left|\Omega(p_n) \cap M_r\right|
}{
    \left|\Omega(p_n)\right|
}.
\label{eq:patch_purity}
\end{equation}
where $\Omega(p_n)$ is the pixel support of patch $p_n$, and $M_r$ is the 
$r$-th segmentation mask, $|\cdot|$ calculates the number of pixels. A patch 
is considered ambiguous if $q_n < \theta$. $\theta=75\%$ is a manually set threshold. 
Third condition stops expansion when no valid unvisited neighbor remains, and the fourth bounds subgraph size and cost.

The sampling process yields $S$ connected patch sets
$\{\{p_{i,j}\}_{j=1}^{N_i}\}_{i=1}^{S}$. 
If represented only as sets, these regions would be analogous to sub-bags and would discard the internal relationships among their constituent patches. 
We therefore consider spatially adjacent patches as graph nodes and connect them with graph edges, yielding $S$ SASGs 
$\{G_i\}_{i=1}^{S}$. 
These SASGs explicitly preserve intra-region adjacency and provide the structural basis for subsequent GNN-based topology encoding.

Each SASG therefore has irregular spatial support, variable extent, and visual coherence with respect to the prior, serving as an approximate candidate semantic unit. A Subgraph State Space Module subsequently performs content-adaptive contextualization to modulate information propagation among SASGs.

\subsection{Subgraph State Space Module}
\label{subsec:sgssm}

The diagnostic relevance of a histological pattern may depend on its co-occurrence and relationships with other tissue structures~\cite{chen_spatial_2024}. 
Subgraph State Space Module (SG-SSM) therefore combines a Mamba-based SSM encoder for efficient long-range contextualization among spatially separated SASGs and a GNN for intra-subgraph topology encoding.

\subsubsection{Long-Range Context among Subgraphs}
The number of SASGs could be enormous due to the large size of WSIs, therefore, a scalable and efficient encoder, Mamba~\cite{gu_mamba_2024}, is incorporated 
to process the batch of SASGs. 
To retain complete slide coverage, each individual patch is additionally treated as a zero-step subgraph.

To provide an ordered input to Mamba, SASGs are first pooled into subgraph tokens 
and are then serialized using a simple spatially informed scan based on the 
coordinates of their seed patches (left-top to bottom-right). 
The ordering is treated as a practical graph-to-sequence transformation rather than as a 
pathological prior. Alternative serialization strategies are evaluated in subsequent ablation study. Let $L=S+N$ denote the total number of multi-patch 
and zero-step subgraphs. For a scan permutation $\pi$,
\begin{equation}
\begin{aligned}
\mathbf{T}_{\pi}
&=
\operatorname{Scan}
\left(
\{\mathbf{t}_i\}_{i=1}^{L},\pi
\right) \\
&=
\left[
\mathbf{t}_{\pi(1)};
\mathbf{t}_{\pi(2)};
\cdots;
\mathbf{t}_{\pi(L)}
\right]
\in \mathbb{R}^{L\times d}.
\end{aligned}
\label{eq:scan}
\end{equation}

The subgraph token $\mathbf{t}_i$ is obtained by pooling its node features:
\begin{equation}
\mathbf{t}_i
=
\operatorname{Pool}(\mathbf{X}_i,V_i)
=
\frac{1}{|V_i|}
\sum_{v\in V_i}\mathbf{x}_{i,v}
\in \mathbb{R}^{d}.
\label{eq:subgraph_pool}
\end{equation}

The SSM encoder is instantiated with a LayerNorm layer~\cite{ba_layer_2016} and a vanilla Mamba block~\cite{gu_mamba_2024}. 
The forward propagation is as follows:
\begin{equation}
\begin{aligned}
\widetilde{\mathbf{T}}_{\pi}
&=
\operatorname{SSMEncoder}
\left(\mathbf{T}_{\pi}\right) \\
&=
\operatorname{MambaBlock}
\left[
\operatorname{LN}
\left(\mathbf{T}_{\pi}\right)
\right]
.
\end{aligned}
\label{eq:ssm_encoder}
\end{equation}

Then recover the contextualized token associated with subgraph $i$ by reversing 
the scan permutation:
\begin{equation}
\widetilde{\mathbf{t}}_i
=
\left[
\pi^{-1}
\left(
\widetilde{\mathbf{T}}_{\pi}
\right)
\right]_i,
\label{eq:inverse_scan}
\end{equation}
where $\widetilde{\mathbf{t}}_i$ denotes the contextualized representation of 
the $i$-th subgraph.

The input-dependent state-space parameters of Mamba provide content-adaptive long-range contextualization~\cite{gu_mamba_2024}.

\subsubsection{Local Topology inside Subgraphs}

Because direct pooling would discard SASG topology, a GNN encoder is therefore employed to propagate information along intra-subgraph edges.

The GNN encoder is instantiated with a Graph Convolution block~\cite{morris_weisfeiler_2019}, a LayerNorm 
layer~\cite{ba_layer_2016}, an activation function and a dropout layer. The forward propagation is 
as follows:
\begin{equation}
\begin{aligned}
\mathbf{H}_i
&=
\operatorname{GNNEncoder}
\left(
\mathbf{X}_i,\mathbf{A}_i
\right) \\
&=
\operatorname{Dropout}
\left(
\sigma
\left(
\operatorname{LN}
\left(
\operatorname{GCN}
\left(
\mathbf{X}_i,\mathbf{A}_i
\right)
\right)
\right)
\right).
\end{aligned}
\label{eq:gnn_encoder}
\end{equation}

The topology-aware subgraph representation is subsequently obtained by pooling:
\begin{equation}
\mathbf{t}_i^{\mathrm{topo}}
=
\operatorname{Pool}
\left(
\mathbf{H}_i,V_i
\right)
=
\frac{1}{|V_i|}
\sum_{v\in V_i}
\mathbf{h}_{i,v}.
\label{eq:topology_pool}
\end{equation}
where $\mathbf{H}_i$ represents the updated node feature matrix of the $i$-th 
SASG, and $\mathbf{t}_i^{\mathrm{topo}}$ denotes the $i$-th subgraph 
representation enhanced with topological information.

\subsubsection{Incorporation of Encoders}

Intuitively, feeding the SASGs into two encoders separately and then adding the 
output features would incorporate two encoders together. We refer this intuitive 
incorporation as a parallel style.

However, two encoders serve relatively independent roles in parallel style. 
We further explore a serial style where the GNN encoder first embeds 
intra-subgraph topology, after which the resulting topology-aware representations 
are tokenized and contextualized by SSM encoder, as illustrated in 
Fig.~\ref{fig:sasg_ssm}(b). Full propagation of the $i$-th SASG is as follows:
\begin{equation}
\begin{aligned}
\mathbf{t}_i^{\mathrm{topo}}
&=
\operatorname{Pool}
\left[
\operatorname{GNNEncoder}
\left(
\mathbf{X}_i,\mathbf{A}_i
\right),
V_i
\right],
\quad i=1,\ldots,L,
\\
\widetilde{\mathbf{T}}_{\pi}^{\mathrm{ser}}
&=
\operatorname{SSMEncoder}
\left[
\operatorname{Scan}
\left(
\{\mathbf{t}_i^{\mathrm{topo}}\}_{i=1}^{L},
\pi
\right)
\right].
\end{aligned}
\label{eq:serial_style}
\end{equation}

The serial design establishes an explicit local-to-global information flow and is adopted as the default configuration due to its superior results.
Detailed comparisons are provided in further analysis of SG-SSM.

\subsection{Classification and Loss Function}
\label{subsec:classification}
Multi-patch SASGs act as intermediate carriers of local structural and long-range 
contextual information. 
Because different SASGs may overlap and repeatedly contain the same patch nodes, directly aggregating all SASG tokens could duplicate evidence from heavily sampled regions. 
We therefore use the contextualized zero-step subgraphs as a unique patch-level readout set and aggregate them using gated attention~\cite{ilse_attention-based_2018}.

Let $Z$ denote the index set of zero-step subgraphs. For each $i\in Z$, 
the unnormalized attention score is computed as:
\begin{equation}
e_i =
\mathbf{W}_{\alpha}
\left[
\tanh
\left(
\mathbf{V}_{\alpha}\widetilde{\mathbf{t}}_i
\right)
\odot
\operatorname{sigm}
\left(
\mathbf{U}_{\alpha}\widetilde{\mathbf{t}}_i
\right)
\right],
\label{eq:attention_score}
\end{equation}
and the normalized attention weights and slide-level representation are 
obtained as:
\begin{equation}
\begin{aligned}
\alpha_i
&=
\frac{\exp(e_i)}
{\displaystyle\sum_{j\in Z}\exp(e_j)},
\quad i\in Z,\quad
\mathbf{z}
=
\sum_{i\in Z}
\alpha_i\widetilde{\mathbf{t}}_i
\in \mathbb{R}^{d}.
\end{aligned}
\label{eq:gated_attention}
\end{equation}
followed by the slide-level prediction:
\begin{equation}
\widehat{y}
=
\operatorname{softmax}
\left(
\operatorname{linear}
\left(
\mathbf{z}
\right)
\right),
\label{eq:prediction}
\end{equation}
where $\widehat{y}$ represents the final prediction label, $\mathbf{z}$ represents 
slide-level feature, $\mathbf{W}_{\alpha}$, $\mathbf{U}_{\alpha}$, and 
$\mathbf{V}_{\alpha}$ represents trainable matrices, $\tanh(\cdot)$ and 
$\operatorname{sigm}(\cdot)$ represents activation functions, and $\odot$ 
represents element-wise multiplication.

For a $C$-class subtyping tasks, the cross entropy (CE) loss is adopted:
\begin{equation}
L_{\mathrm{CE}}
=
-
\sum_{c=1}^{C}
y_c
\log
\left(
\widehat{y}_c
\right).
\label{eq:ce_loss}
\end{equation}

\section{Experiments}
\label{sec:experiments}

\subsection{Datasets}
\label{subsec:datasets}

To validate the effectiveness of SASG-SSM for WSI subtyping, we conduct 
extensive experiments on four cohorts from The Cancer Genome Atlas (TCGA), with details summarized below.

\textit{ESCA:} It comprises two subtypes of esophageal carcinoma: squamous 
cell carcinoma (SCC, 92 slides from 90 patients) and adenocarcinoma 
(AC, 66 slides from 66 patients), totaling 158 diagnostic WSIs.

\textit{BRCA:} It encompasses the major subtypes of invasive breast carcinoma. 
In this study, we focus on two predominant subtypes: invasive lobular carcinoma 
(ILC, 188 slides from 175 patients) and invasive ductal carcinoma 
(IDC, 769 slides from 722 patients), totaling 957 diagnostic WSIs.

\textit{NSCLC:} It consists of two subtypes of lung cancer: lung adenocarcinoma 
(LUAD, 489 slides from 430 cases) and lung squamous cell carcinoma 
(LUSC, 512 slides from 478 cases), totaling 1,001 diagnostic WSIs.

\textit{RCC:} It covers three subtypes: kidney chromophobe renal cell carcinoma 
(CHRCC, 105 slides from 95 cases), kidney clear cell renal cell carcinoma 
(CCRCC, 435 slides from 429 cases), and kidney papillary renal cell carcinoma 
(PRCC, 255 slides from 234 cases), totaling 795 diagnostic WSIs.

We additionally downsample BRCA, NSCLC, and RCC training sets to scales comparable to ESCA while preserving class proportions, enabling evaluation under limited data availability. Details are summarized in Table~\ref{tab:datasets}.

\begin{table}[!h]
\caption{Summarized Details of Datasets and Training Splits}
\label{tab:datasets}
\centering
\setlength{\tabcolsep}{2.5pt}
\renewcommand{\arraystretch}{1.1}

\resizebox{0.5\columnwidth}{!}{%
\begin{tabular}{l|cc|cc|cc|ccc}
\hline
\multirow{2}{*}{Item}
& \multicolumn{2}{c|}{ESCA}
& \multicolumn{2}{c|}{BRCA}
& \multicolumn{2}{c|}{NSCLC}
& \multicolumn{3}{c}{RCC} \\
\cline{2-10}
& SCC & AC
& ILC & IDC
& LUAD & LUSC
& CHRCC & CCRCC & PRCC \\
\hline
Class label
& 1 & 0
& 1 & 0
& 1 & 0
& 2 & 1 & 0 \\

No. of slides
& 92 & 66
& 188 & 769
& 489 & 512
& 105 & 435 & 255 \\

Train-small
& 72 & 54
& 24 & 93
& 58 & 60
& 19 & 66 & 39 \\

Train-large
& -- & --
& 150 & 631
& 406 & 395
& 84 & 350 & 207 \\

Validation
& 20 & 12
& 38 & 138
& 83 & 117
& 21 & 85 & 48 \\
\hline
\end{tabular}%
}
\end{table}

\subsection{Implementation Details}
\label{subsec:implementation}

\textit{Preprocessing:}
Following~\cite{lu_data-efficient_2021}, foreground tissue is segmented and cropped into non-overlapping $256\times256$ patches at $20\times$ magnification, used for all image operations unless specified otherwise. 
Because this rigid grid is agnostic to tissue boundaries, SAM2~\cite{ravi_sam_2024} masks provide structure-aware semantic priors. 
WSIs are first divided into $2560\times2560$ tiles with 256-pixel overlap to reduce boundary discontinuities.
Each tile spans $10\times10$ patches and provides local context. 
The resulting masks guide patch grouping into spatially coherent candidate semantic units.


\textit{Segmentation Settings:}
Since SAM2 is primarily pretrained on natural images, we use point prompts 
to guide its segmentation of histopathological structures. Specifically, 
point prompts are sampled from regions obtained by threshold-based partitioning, 
following the empirical observation that tumor and stromal regions often 
differ in staining intensity.

\textit{Training Settings:}
SASG-SSM and all baseline methods are implemented using PyTorch and trained 
on eight NVIDIA A6000 GPUs. The Adam optimizer~\cite{kingma_adam_2015} is used with a learning rate 
of $2\times10^{-5}$ and a weight decay of $1\times10^{-5}$. The batch size 
is set to 1, with a maximum of 200 training epochs and an early-stopping 
strategy.

\textit{Evaluation:}
Performance is assessed using the F1 score (F1), area under the receiver 
operating characteristic curve (AUC), and accuracy (ACC). To evaluate model robustness, we performed fivefold cross-validation using patient-level splits.

\subsection{Comparisons with State-of-the-Art}
\label{subsec:sota}

We conduct a comprehensive comparison against representative state-of-the-art  
(SOTA) methods for WSI subtyping. The compared methods are divided into three  
categories: four MIL-based methods, including ABMIL~\cite{ilse_attention-based_2018},  
TransMIL~\cite{shao_transmil_2021}, ILRA~\cite{xiang_exploring_2022}, and ACMIL~\cite{zhang_attention-challenging_2025}; two GNN-based  
methods, including PatchGCN~\cite{chen_whole_2021} and SGMF~\cite{shi_structure-aware_2023}; and two SSM-based  
methods, including MambaMIL~\cite{yang_mambamil_2024} and PAM~\cite{huang_unleash_2025}. To ensure a fair  
comparison, all baseline methods are trained and evaluated using the same fivefold cross-validation splits as SASG-SSM. 
We follow the hyperparameter  
settings and reproduction strategies reported in the original studies, with  
the exception of the feature encoder, which is standardized as an  
ImageNet-pretrained~\cite{deng_imagenet_2009} ResNet50~\cite{he_deep_2016} with a 1024-dimensional output for all methods. 

\subsubsection{Overall Comparison}
As shown in Table~\ref{tab:comparison}, SASG-SSM achieves the highest F1 scores in both small- and large-set settings. 
In the small-set setting, it also achieves the best F1 and AUC across all datasets.
Notably, on TCGA-ESCA, it reaches 92.17\% F1, 95.71\% AUC, and 91.09\% ACC, exceeding the second-best results by 2.93, 0.71, and 3.87 points, respectively. 
In the large-set setting, it retains the highest F1 and competitive performance on other metrics, indicating balanced classification across histopathological subtypes.

\begin{table*}[!h]
\caption{Overall Comparison with State-of-the-Art Methods on Four TCGA Datasets}
\label{tab:comparison}
\centering
\setlength{\tabcolsep}{1.5pt}
\renewcommand{\arraystretch}{1.05}

\resizebox{\textwidth}{!}{%
\begin{tabular}{l|ccc|ccc|ccc|ccc}
\hline
\multirow{2}{*}[-1pt]{Method}
& \multicolumn{3}{c|}{ESCA}
& \multicolumn{3}{c|}{BRCA}
& \multicolumn{3}{c|}{NSCLC}
& \multicolumn{3}{c}{RCC} \\

\noalign{\vskip-\aboverulesep}
\cmidrule(l{3pt}r{3pt}){2-4}
\cmidrule(l{3pt}r{3pt}){5-7}
\cmidrule(l{3pt}r{3pt}){8-10}
\cmidrule(l{3pt}r{3pt}){11-13}
\noalign{\vskip-\belowrulesep}

& F1 & AUC & ACC
& F1 & AUC & ACC
& F1 & AUC & ACC
& F1 & AUC & ACC \\
\hline

\multicolumn{13}{c}{Small Training Set ($n \approx 120$)} \\
\hline

ABMIL
& 57.24 {\tiny $\pm$29.76} & 78.79 {\tiny $\pm$10.43} & 61.13 {\tiny $\pm$19.16}
& 10.96 {\tiny $\pm$21.92} & 74.30 {\tiny $\pm$3.60} & 80.75 {\tiny $\pm$1.51}
& 60.08 {\tiny $\pm$30.34} & 82.27 {\tiny $\pm$5.12} & 72.67 {\tiny $\pm$8.84}
& 75.09 {\tiny $\pm$9.41} & 91.68 {\tiny $\pm$1.28} & 81.64 {\tiny $\pm$4.40} \\

TransMIL
& 66.97 {\tiny $\pm$30.71} & 83.11 {\tiny $\pm$19.95} & 72.04 {\tiny $\pm$21.05}
& 9.46 {\tiny $\pm$13.30} & 62.57 {\tiny $\pm$7.08} & 80.42 {\tiny $\pm$1.51}
& 37.31 {\tiny $\pm$25.86} & 65.00 {\tiny $\pm$9.02} & 56.51 {\tiny $\pm$7.35}
& 71.28 {\tiny $\pm$6.78} & 91.06 {\tiny $\pm$3.66} & 79.29 {\tiny $\pm$4.94} \\

ILRA
& 76.35 {\tiny $\pm$9.27} & 87.12 {\tiny $\pm$8.48} & 65.22 {\tiny $\pm$15.57}
& 28.80 {\tiny $\pm$23.88} & 77.34 {\tiny $\pm$3.23} & 81.08 {\tiny $\pm$1.24}
& 72.65 {\tiny $\pm$7.54} & 83.28 {\tiny $\pm$3.35} & 73.23 {\tiny $\pm$5.03}
& 79.09 {\tiny $\pm$3.43} & 93.52 {\tiny $\pm$1.44} & \textbf{83.31} {\tiny $\pm$3.50} \\

ACMIL
& 73.34 {\tiny $\pm$6.01} & 85.51 {\tiny $\pm$5.68} & 58.27 {\tiny $\pm$7.69}
& 0.00 {\tiny $\pm$0.00} & 60.42 {\tiny $\pm$7.03} & 80.32 {\tiny $\pm$1.25}
& 35.24 {\tiny $\pm$28.88} & 71.24 {\tiny $\pm$9.51} & 50.28 {\tiny $\pm$8.02}
& 45.16 {\tiny $\pm$10.85} & 87.15 {\tiny $\pm$4.64} & 69.18 {\tiny $\pm$7.07} \\

PatchGCN
& 83.98 {\tiny $\pm$3.97} & 88.62 {\tiny $\pm$2.39} & 80.85 {\tiny $\pm$5.37}
& 21.79 {\tiny $\pm$24.42} & 72.55 {\tiny $\pm$7.91} & 79.27 {\tiny $\pm$3.34}
& 71.92 {\tiny $\pm$4.64} & 83.36 {\tiny $\pm$2.46} & 73.53 {\tiny $\pm$2.63}
& 80.80 {\tiny $\pm$5.39} & 92.66 {\tiny $\pm$3.37} & 80.60 {\tiny $\pm$5.39} \\

SGMF
& 89.24 {\tiny $\pm$4.02} & 94.12 {\tiny $\pm$3.05} & 87.20 {\tiny $\pm$4.60}
& 26.54 {\tiny $\pm$20.51} & 71.71 {\tiny $\pm$10.68} & \textbf{81.18} {\tiny $\pm$0.89}
& 60.88 {\tiny $\pm$17.46} & 83.68 {\tiny $\pm$3.83} & 68.77 {\tiny $\pm$6.34}
& 80.00 {\tiny $\pm$4.67} & 93.84 {\tiny $\pm$1.19} & 80.11 {\tiny $\pm$4.47} \\

MambaMIL
& 89.08 {\tiny $\pm$3.74} & 92.98 {\tiny $\pm$4.12} & 87.22 {\tiny $\pm$4.67}
& 37.43 {\tiny $\pm$22.98} & 74.57 {\tiny $\pm$7.27} & 67.35 {\tiny $\pm$22.55}
& 72.29 {\tiny $\pm$9.50} & 83.23 {\tiny $\pm$3.45} & 74.99 {\tiny $\pm$5.72}
& 81.10 {\tiny $\pm$7.99} & 92.80 {\tiny $\pm$5.61} & 81.15 {\tiny $\pm$7.88} \\

PAM
& 88.69 {\tiny $\pm$5.95} & 95.00 {\tiny $\pm$2.13} & 86.20 {\tiny $\pm$6.99}
& 38.38 {\tiny $\pm$18.36} & 76.00 {\tiny $\pm$7.97} & 80.62 {\tiny $\pm$1.91}
& 62.45 {\tiny $\pm$17.85} & 82.47 {\tiny $\pm$4.01} & 69.65 {\tiny $\pm$9.24}
& 80.03 {\tiny $\pm$8.48} & 93.52 {\tiny $\pm$3.03} & 80.04 {\tiny $\pm$8.34} \\

\textbf{SASG-SSM}
& \textbf{92.17} {\tiny $\pm$4.16} & \textbf{95.71} {\tiny $\pm$3.12} & \textbf{91.09} {\tiny $\pm$4.61}
& \textbf{38.82} {\tiny $\pm$16.59} & \textbf{77.53} {\tiny $\pm$5.08} & 80.14 {\tiny $\pm$1.75}
& \textbf{75.15} {\tiny $\pm$1.82} & \textbf{84.87} {\tiny $\pm$2.81} & \textbf{76.52} {\tiny $\pm$1.87}
& \textbf{81.37} {\tiny $\pm$3.52} & \textbf{93.85} {\tiny $\pm$1.77} & 81.17 {\tiny $\pm$3.33} \\

\hline
\multicolumn{13}{c}{Large Traning Set ($n \approx 1000$)} \\
\hline

ABMIL
& -- & -- & --
& 59.11 {\tiny $\pm$10.25} & 87.36 {\tiny $\pm$4.10} & 85.18 {\tiny $\pm$2.67}
& 81.63 {\tiny $\pm$8.50} & 92.73 {\tiny $\pm$1.36} & 83.43 {\tiny $\pm$4.34}
& 85.46 {\tiny $\pm$3.42} & 96.41 {\tiny $\pm$1.20} & 88.30 {\tiny $\pm$2.88} \\

TransMIL
& -- & -- & --
& 55.70 {\tiny $\pm$7.00} & 84.81 {\tiny $\pm$2.34} & 83.26 {\tiny $\pm$1.27}
& 81.88 {\tiny $\pm$4.34} & 90.74 {\tiny $\pm$1.89} & 83.33 {\tiny $\pm$2.91}
& 85.90 {\tiny $\pm$2.33} & 96.55 {\tiny $\pm$1.62} & 89.19 {\tiny $\pm$1.80} \\

ILRA
& -- & -- & --
& 63.12 {\tiny $\pm$2.14} & 87.18 {\tiny $\pm$1.99} & 86.86 {\tiny $\pm$2.25}
& 84.12 {\tiny $\pm$2.55} & 92.65 {\tiny $\pm$1.91} & 84.91 {\tiny $\pm$1.68}
& 85.65 {\tiny $\pm$1.91} & 97.46 {\tiny $\pm$1.36} & 89.06 {\tiny $\pm$2.68} \\

ACMIL
& -- & -- & --
& 47.72 {\tiny $\pm$25.60} & 87.52 {\tiny $\pm$1.69} & 85.80 {\tiny $\pm$3.53}
& 80.73 {\tiny $\pm$4.86} & 91.59 {\tiny $\pm$3.08} & 82.92 {\tiny $\pm$3.12}
& 85.96 {\tiny $\pm$4.41} & \textbf{97.69} {\tiny $\pm$0.91} & 89.09 {\tiny $\pm$3.83} \\

PatchGCN
& -- & -- & --
& 61.04 {\tiny $\pm$10.10} & 88.10 {\tiny $\pm$3.21} & 87.62 {\tiny $\pm$1.98}
& 85.67 {\tiny $\pm$2.25} & 93.04 {\tiny $\pm$1.92} & 85.92 {\tiny $\pm$2.01}
& 88.83 {\tiny $\pm$2.46} & 97.57 {\tiny $\pm$0.91} & 88.82 {\tiny $\pm$2.43} \\

SGMF
& -- & -- & --
& 64.46 {\tiny $\pm$5.94} & 88.22 {\tiny $\pm$1.77} & 87.43 {\tiny $\pm$1.82}
& 84.99 {\tiny $\pm$3.95} & 92.68 {\tiny $\pm$1.66} & 85.40 {\tiny $\pm$3.54}
& 87.76 {\tiny $\pm$5.83} & 97.52 {\tiny $\pm$1.32} & 87.53 {\tiny $\pm$6.30} \\

MambaMIL
& -- & -- & --
& 65.15 {\tiny $\pm$5.83} & 88.31 {\tiny $\pm$3.20} & 86.92 {\tiny $\pm$1.14}
& 85.53 {\tiny $\pm$2.61} & 92.99 {\tiny $\pm$1.66} & 85.62 {\tiny $\pm$3.25}
& 89.18 {\tiny $\pm$2.52} & 97.56 {\tiny $\pm$0.89} & 89.15 {\tiny $\pm$2.60} \\

PAM
& -- & -- & --
& 59.85 {\tiny $\pm$9.48} & 87.74 {\tiny $\pm$4.86} & 86.41 {\tiny $\pm$1.57}
& 85.52 {\tiny $\pm$2.13} & 92.97 {\tiny $\pm$0.94} & 85.58 {\tiny $\pm$1.38}
& 88.08 {\tiny $\pm$3.27} & 97.55 {\tiny $\pm$0.91} & 88.17 {\tiny $\pm$3.17} \\

\textbf{SASG-SSM}
& -- & -- & --
& \textbf{65.64} {\tiny $\pm$5.37} & \textbf{88.37} {\tiny $\pm$3.20} & \textbf{88.06} {\tiny $\pm$1.07}
& \textbf{86.82} {\tiny $\pm$1.53} & \textbf{93.06} {\tiny $\pm$0.81} & \textbf{87.01} {\tiny $\pm$1.78}
& \textbf{89.85} {\tiny $\pm$2.95} & 97.67 {\tiny $\pm$1.22} & \textbf{89.84} {\tiny $\pm$3.05} \\

\hline
\end{tabular}%
}

\end{table*}


\subsubsection{Few-Shot Comparison}

To further evaluate SASG-SSM under more restricted data availability, we 
conduct few-shot training by randomly sampling varying numbers of slides from 
the fivefold training splits while retaining the same validation splits. As 
shown in Table~\ref{tab:few_shot}, SASG-SSM consistently outperforms the 
competing methods, achieving the strongest overall F1 and AUC performance across the evaluated settings
These results indicate that the proposed Semantic-Aware Subgraphs can extract 
discriminative histopathological representations under 
extremely limited supervision.

\begin{table*}[!h]
\caption{
Few-shot Comparison with State-of-the-Art Methods on Four TCGA Datasets. ($k$ Denotes the Number of Slides per Class.)}
\label{tab:few_shot}
\centering
\setlength{\tabcolsep}{2pt}
\renewcommand{\arraystretch}{1.05}
\resizebox{\textwidth}{!}{%
\begin{tabular}{ll|cc|cc|cc|cc|cc}
\hline
\multirow{2}{*}{Dataset} & \multirow{2}{*}{Method}
& \multicolumn{2}{c|}{$k=1$}
& \multicolumn{2}{c|}{$k=2$}
& \multicolumn{2}{c|}{$k=4$}
& \multicolumn{2}{c|}{$k=8$}
& \multicolumn{2}{c}{$k=16$} \\
\noalign{\vskip-\aboverulesep}
\cmidrule(l{3pt}r{3pt}){3-4}
\cmidrule(l{3pt}r{3pt}){5-6}
\cmidrule(l{3pt}r{3pt}){7-8}
\cmidrule(l{3pt}r{3pt}){9-10}
\cmidrule(l{3pt}r{3pt}){11-12}
\noalign{\vskip-\belowrulesep}
& & F1 & AUC & F1 & AUC & F1 & AUC & F1 & AUC & F1 & AUC \\
\hline
\multirow{9}{*}{ESCA} & ABMIL & 57.91 {\tiny $\pm$29.87} & 60.73 {\tiny $\pm$9.70} & 44.63 {\tiny $\pm$36.81} & 56.81 {\tiny $\pm$13.42} & 58.94 {\tiny $\pm$30.07} & 79.93 {\tiny $\pm$5.90} & 46.17 {\tiny $\pm$33.98} & 82.63 {\tiny $\pm$5.50} & 58.98 {\tiny $\pm$30.24} & 82.13 {\tiny $\pm$6.92} \\
& TransMIL & 59.56 {\tiny $\pm$30.25} & 55.36 {\tiny $\pm$16.20} & 43.92 {\tiny $\pm$24.52} & 63.57 {\tiny $\pm$12.90} & 52.85 {\tiny $\pm$30.32} & 60.63 {\tiny $\pm$13.46} & 60.81 {\tiny $\pm$13.55} & 63.44 {\tiny $\pm$11.35} & 52.33 {\tiny $\pm$26.81} & 73.52 {\tiny $\pm$6.23} \\
& ILRA & 58.94 {\tiny $\pm$30.07} & 60.58 {\tiny $\pm$13.09} & 57.95 {\tiny $\pm$29.54} & 71.58 {\tiny $\pm$13.40} & 58.79 {\tiny $\pm$29.98} & 79.85 {\tiny $\pm$8.03} & 56.09 {\tiny $\pm$28.16} & 81.81 {\tiny $\pm$4.49} & 73.34 {\tiny $\pm$6.01} & 83.01 {\tiny $\pm$6.47} \\
& ACMIL & 57.73 {\tiny $\pm$29.85} & 59.88 {\tiny $\pm$13.66} & 59.23 {\tiny $\pm$30.16} & 72.06 {\tiny $\pm$14.56} & 43.55 {\tiny $\pm$36.02} & 76.30 {\tiny $\pm$11.43} & 58.94 {\tiny $\pm$30.07} & 84.38 {\tiny $\pm$3.34} & 71.36 {\tiny $\pm$10.05} & 83.04 {\tiny $\pm$3.98} \\
& PatchGCN & 74.13 {\tiny $\pm$5.33} & 76.29 {\tiny $\pm$9.35} & 70.67 {\tiny $\pm$7.80} & 72.10 {\tiny $\pm$14.17} & 64.37 {\tiny $\pm$11.54} & 78.84 {\tiny $\pm$7.78} & 69.06 {\tiny $\pm$10.16} & 77.47 {\tiny $\pm$7.41} & 77.77 {\tiny $\pm$6.95} & 80.55 {\tiny $\pm$9.03} \\
& SGMF & 72.53 {\tiny $\pm$5.95} & 55.43 {\tiny $\pm$13.22} & 70.12 {\tiny $\pm$14.06} & 70.38 {\tiny $\pm$18.98} & 56.47 {\tiny $\pm$26.24} & 78.23 {\tiny $\pm$7.26} & 69.16 {\tiny $\pm$12.36} & 78.49 {\tiny $\pm$10.12} & 63.00 {\tiny $\pm$29.16} & 83.35 {\tiny $\pm$5.39} \\
& MambaMIL & 74.92 {\tiny $\pm$6.92} & 73.31 {\tiny $\pm$5.35} & 69.14 {\tiny $\pm$7.79} & 68.26 {\tiny $\pm$14.49} & 72.67 {\tiny $\pm$5.30} & 74.27 {\tiny $\pm$14.64} & 69.10 {\tiny $\pm$7.87} & 79.54 {\tiny $\pm$7.35} & 74.99 {\tiny $\pm$10.86} & 79.25 {\tiny $\pm$11.11} \\
& PAM & 54.41 {\tiny $\pm$19.24} & 59.35 {\tiny $\pm$11.07} & 71.79 {\tiny $\pm$5.04} & 63.05 {\tiny $\pm$11.32} & 71.21 {\tiny $\pm$4.41} & 66.98 {\tiny $\pm$8.60} & 72.49 {\tiny $\pm$3.97} & 76.54 {\tiny $\pm$7.79} & 75.73 {\tiny $\pm$7.13} & 81.74 {\tiny $\pm$7.42} \\
& \textbf{SASG-SSM} & \textbf{75.07} {\tiny $\pm$5.65} & \textbf{76.75} {\tiny $\pm$1.77} & \textbf{72.45} {\tiny $\pm$9.54} & \textbf{78.99} {\tiny $\pm$10.75} & \textbf{72.82} {\tiny $\pm$6.93} & \textbf{82.26} {\tiny $\pm$2.75} & \textbf{73.86} {\tiny $\pm$5.94} & \textbf{84.81} {\tiny $\pm$3.28} & \textbf{79.00} {\tiny $\pm$9.10} & \textbf{86.78} {\tiny $\pm$5.80} \\
\hline
\multirow{9}{*}{BRCA} & ABMIL & 20.88 {\tiny $\pm$14.52} & 43.91 {\tiny $\pm$6.86} & 21.42 {\tiny $\pm$14.34} & 50.80 {\tiny $\pm$7.50} & \textbf{32.96} {\tiny $\pm$2.07} & 58.39 {\tiny $\pm$6.48} & 30.90 {\tiny $\pm$5.28} & 53.99 {\tiny $\pm$7.49} & 16.34 {\tiny $\pm$13.86} & 69.66 {\tiny $\pm$5.01} \\
& TransMIL & 14.61 {\tiny $\pm$15.96} & 51.76 {\tiny $\pm$11.01} & 24.33 {\tiny $\pm$12.60} & 54.40 {\tiny $\pm$7.25} & 18.30 {\tiny $\pm$16.09} & 56.14 {\tiny $\pm$11.01} & 18.97 {\tiny $\pm$15.80} & 54.95 {\tiny $\pm$7.40} & 30.04 {\tiny $\pm$4.18} & 54.89 {\tiny $\pm$7.63} \\
& ILRA & 7.10 {\tiny $\pm$14.21} & 52.11 {\tiny $\pm$8.95} & 19.82 {\tiny $\pm$16.27} & 55.58 {\tiny $\pm$9.58} & 26.29 {\tiny $\pm$13.26} & 58.86 {\tiny $\pm$9.66} & 26.17 {\tiny $\pm$13.20} & 57.44 {\tiny $\pm$11.32} & 7.07 {\tiny $\pm$14.14} & 70.02 {\tiny $\pm$3.61} \\
& ACMIL & 22.09 {\tiny $\pm$14.16} & 48.19 {\tiny $\pm$11.83} & 18.56 {\tiny $\pm$15.30} & 56.59 {\tiny $\pm$5.26} & 20.62 {\tiny $\pm$14.76} & 60.19 {\tiny $\pm$8.21} & 26.83 {\tiny $\pm$13.51} & 56.51 {\tiny $\pm$11.36} & 11.15 {\tiny $\pm$12.85} & 67.63 {\tiny $\pm$4.48} \\
& PatchGCN & 1.58 {\tiny $\pm$1.87} & 55.00 {\tiny $\pm$11.87} & 6.40 {\tiny $\pm$9.71} & 54.17 {\tiny $\pm$8.66} & 7.97 {\tiny $\pm$6.44} & 58.32 {\tiny $\pm$8.10} & 7.90 {\tiny $\pm$14.43} & 55.57 {\tiny $\pm$10.83} & 18.51 {\tiny $\pm$17.32} & 64.22 {\tiny $\pm$7.38} \\
& SGMF & 22.53 {\tiny $\pm$12.15} & 48.88 {\tiny $\pm$6.38} & 24.77 {\tiny $\pm$12.22} & 55.13 {\tiny $\pm$5.18} & 30.72 {\tiny $\pm$3.85} & 59.10 {\tiny $\pm$3.32} & 29.83 {\tiny $\pm$13.88} & 60.29 {\tiny $\pm$12.19} & 22.00 {\tiny $\pm$16.47} & 67.75 {\tiny $\pm$6.91} \\
& MambaMIL & 2.00 {\tiny $\pm$2.74} & 49.07 {\tiny $\pm$10.69} & 2.88 {\tiny $\pm$4.36} & 53.36 {\tiny $\pm$11.89} & 3.75 {\tiny $\pm$8.39} & 59.77 {\tiny $\pm$5.32} & 8.63 {\tiny $\pm$19.29} & 58.73 {\tiny $\pm$13.86} & 11.04 {\tiny $\pm$18.05} & 64.93 {\tiny $\pm$12.35} \\
& PAM & 4.88 {\tiny $\pm$7.48} & 52.94 {\tiny $\pm$11.50} & 6.37 {\tiny $\pm$6.00} & 53.70 {\tiny $\pm$8.77} & 15.60 {\tiny $\pm$12.92} & 56.78 {\tiny $\pm$5.72} & 22.28 {\tiny $\pm$10.48} & 56.27 {\tiny $\pm$7.16} & 18.28 {\tiny $\pm$15.92} & 63.96 {\tiny $\pm$1.90} \\
& \textbf{SASG-SSM} & \textbf{22.76} {\tiny $\pm$12.82} & \textbf{57.43} {\tiny $\pm$9.79} & \textbf{25.27} {\tiny $\pm$5.83} & \textbf{59.39} {\tiny $\pm$5.44} & 32.92 {\tiny $\pm$12.70} & \textbf{63.28} {\tiny $\pm$6.05} & \textbf{38.96} {\tiny $\pm$4.61} & \textbf{65.73} {\tiny $\pm$5.09} & \textbf{38.47} {\tiny $\pm$6.66} & \textbf{70.64} {\tiny $\pm$5.10} \\
\hline
\multirow{9}{*}{NSCLC} & ABMIL & 38.54 {\tiny $\pm$31.78} & 59.12 {\tiny $\pm$6.87} & 27.28 {\tiny $\pm$33.42} & 63.30 {\tiny $\pm$3.61} & 51.32 {\tiny $\pm$26.06} & 66.81 {\tiny $\pm$2.65} & 45.56 {\tiny $\pm$21.95} & 64.60 {\tiny $\pm$3.49} & 34.55 {\tiny $\pm$29.22} & 70.73 {\tiny $\pm$3.62} \\
& TransMIL & 11.54 {\tiny $\pm$15.45} & 55.50 {\tiny $\pm$7.64} & 25.68 {\tiny $\pm$22.93} & 60.62 {\tiny $\pm$3.35} & 31.73 {\tiny $\pm$26.47} & 62.09 {\tiny $\pm$2.31} & 34.87 {\tiny $\pm$29.13} & 58.02 {\tiny $\pm$10.34} & 21.26 {\tiny $\pm$21.95} & 62.01 {\tiny $\pm$5.04} \\
& ILRA & 37.83 {\tiny $\pm$31.09} & 63.23 {\tiny $\pm$3.63} & 27.28 {\tiny $\pm$33.42} & 62.58 {\tiny $\pm$2.72} & 25.15 {\tiny $\pm$30.92} & 64.04 {\tiny $\pm$4.62} & 39.89 {\tiny $\pm$32.73} & 60.44 {\tiny $\pm$4.35} & 61.56 {\tiny $\pm$7.22} & 69.07 {\tiny $\pm$3.57} \\
& ACMIL & 53.50 {\tiny $\pm$25.05} & 54.06 {\tiny $\pm$7.33} & 37.32 {\tiny $\pm$31.25} & 62.85 {\tiny $\pm$2.12} & 41.83 {\tiny $\pm$23.60} & 64.29 {\tiny $\pm$3.69} & 53.96 {\tiny $\pm$20.89} & 63.96 {\tiny $\pm$2.94} & 40.96 {\tiny $\pm$28.52} & 68.82 {\tiny $\pm$3.76} \\
& PatchGCN & 55.63 {\tiny $\pm$8.93} & 59.18 {\tiny $\pm$6.35} & 46.08 {\tiny $\pm$15.29} & 61.57 {\tiny $\pm$3.96} & 37.81 {\tiny $\pm$24.28} & 65.09 {\tiny $\pm$7.08} & 60.66 {\tiny $\pm$8.82} & 62.12 {\tiny $\pm$5.34} & 57.09 {\tiny $\pm$8.52} & 65.31 {\tiny $\pm$10.69} \\
& SGMF & 38.14 {\tiny $\pm$28.57} & 58.05 {\tiny $\pm$6.39} & 56.64 {\tiny $\pm$7.21} & 60.93 {\tiny $\pm$2.23} & 60.01 {\tiny $\pm$6.90} & 64.45 {\tiny $\pm$4.19} & 64.04 {\tiny $\pm$6.88} & 60.28 {\tiny $\pm$6.28} & 65.21 {\tiny $\pm$4.85} & 70.34 {\tiny $\pm$3.53} \\
& MambaMIL & 47.81 {\tiny $\pm$3.17} & 57.84 {\tiny $\pm$7.94} & 50.42 {\tiny $\pm$13.92} & 61.70 {\tiny $\pm$3.62} & 52.13 {\tiny $\pm$19.60} & 63.43 {\tiny $\pm$2.40} & 45.86 {\tiny $\pm$12.58} & 57.67 {\tiny $\pm$9.23} & 52.83 {\tiny $\pm$24.63} & 70.93 {\tiny $\pm$6.79} \\
& PAM & 49.78 {\tiny $\pm$7.96} & 57.34 {\tiny $\pm$5.12} & 56.64 {\tiny $\pm$9.21} & 57.92 {\tiny $\pm$4.88} & 43.48 {\tiny $\pm$16.68} & 63.22 {\tiny $\pm$4.67} & 61.05 {\tiny $\pm$5.95} & 58.18 {\tiny $\pm$3.07} & 53.88 {\tiny $\pm$8.64} & 64.39 {\tiny $\pm$4.08} \\
& \textbf{SASG-SSM} & \textbf{64.71} {\tiny $\pm$4.56} & \textbf{63.64} {\tiny $\pm$4.18} & \textbf{64.64} {\tiny $\pm$4.96} & \textbf{65.31} {\tiny $\pm$5.06} & \textbf{62.85} {\tiny $\pm$6.50} & \textbf{67.42} {\tiny $\pm$3.20} & \textbf{66.20} {\tiny $\pm$4.79} & \textbf{68.07} {\tiny $\pm$3.85} & \textbf{67.23} {\tiny $\pm$4.43} & \textbf{71.14} {\tiny $\pm$3.53} \\
\hline
\multirow{9}{*}{RCC} & ABMIL & 27.14 {\tiny $\pm$16.60} & 69.91 {\tiny $\pm$8.33} & 14.49 {\tiny $\pm$6.08} & 74.61 {\tiny $\pm$6.42} & 44.40 {\tiny $\pm$19.82} & 82.48 {\tiny $\pm$3.07} & 35.85 {\tiny $\pm$18.27} & 84.49 {\tiny $\pm$3.84} & 44.70 {\tiny $\pm$28.56} & 84.34 {\tiny $\pm$4.68} \\
& TransMIL & 38.49 {\tiny $\pm$16.32} & 72.47 {\tiny $\pm$8.06} & 25.08 {\tiny $\pm$13.68} & 70.96 {\tiny $\pm$7.81} & 50.35 {\tiny $\pm$7.03} & 78.99 {\tiny $\pm$6.20} & 61.06 {\tiny $\pm$15.52} & 83.12 {\tiny $\pm$8.05} & 67.39 {\tiny $\pm$9.07} & 85.82 {\tiny $\pm$7.40} \\
& ILRA & 19.20 {\tiny $\pm$6.15} & 69.20 {\tiny $\pm$9.13} & 21.53 {\tiny $\pm$8.80} & 70.87 {\tiny $\pm$12.07} & 17.39 {\tiny $\pm$12.28} & 80.22 {\tiny $\pm$2.83} & 52.70 {\tiny $\pm$18.22} & 84.11 {\tiny $\pm$2.53} & 60.49 {\tiny $\pm$19.19} & 89.61 {\tiny $\pm$1.22} \\
& ACMIL & 41.21 {\tiny $\pm$10.14} & 70.41 {\tiny $\pm$7.67} & 23.36 {\tiny $\pm$7.48} & 74.55 {\tiny $\pm$7.07} & 30.78 {\tiny $\pm$13.65} & 80.87 {\tiny $\pm$2.36} & 35.34 {\tiny $\pm$19.16} & 83.69 {\tiny $\pm$4.23} & 42.40 {\tiny $\pm$19.60} & 84.87 {\tiny $\pm$3.47} \\
& PatchGCN & 47.88 {\tiny $\pm$9.58} & 69.06 {\tiny $\pm$6.72} & 56.92 {\tiny $\pm$5.17} & 73.64 {\tiny $\pm$5.90} & 67.13 {\tiny $\pm$6.34} & 78.74 {\tiny $\pm$3.47} & 65.61 {\tiny $\pm$4.11} & 80.23 {\tiny $\pm$5.46} & 61.49 {\tiny $\pm$12.84} & 82.57 {\tiny $\pm$5.11} \\
& SGMF & 33.35 {\tiny $\pm$14.27} & 69.12 {\tiny $\pm$9.69} & 53.65 {\tiny $\pm$11.07} & 77.09 {\tiny $\pm$6.79} & 64.43 {\tiny $\pm$11.40} & 83.57 {\tiny $\pm$4.60} & 66.81 {\tiny $\pm$7.74} & 87.30 {\tiny $\pm$4.88} & 65.78 {\tiny $\pm$19.97} & 89.75 {\tiny $\pm$3.63} \\
& MambaMIL & 46.68 {\tiny $\pm$11.13} & 63.71 {\tiny $\pm$7.65} & 47.28 {\tiny $\pm$21.22} & 72.28 {\tiny $\pm$11.18} & 59.64 {\tiny $\pm$11.90} & 82.88 {\tiny $\pm$4.67} & 64.72 {\tiny $\pm$10.23} & 85.40 {\tiny $\pm$4.50} & 68.99 {\tiny $\pm$13.24} & 89.57 {\tiny $\pm$3.27} \\
& PAM & 39.85 {\tiny $\pm$3.06} & 65.74 {\tiny $\pm$6.54} & 45.91 {\tiny $\pm$9.04} & 71.62 {\tiny $\pm$10.32} & 56.25 {\tiny $\pm$13.73} & 84.31 {\tiny $\pm$6.32} & 69.71 {\tiny $\pm$2.94} & 87.29 {\tiny $\pm$4.25} & 76.91 {\tiny $\pm$5.35} & 90.14 {\tiny $\pm$5.38} \\
& \textbf{SASG-SSM} & \textbf{57.49} {\tiny $\pm$8.82} & \textbf{74.39} {\tiny $\pm$6.56} & \textbf{62.40} {\tiny $\pm$5.98} & \textbf{77.90} {\tiny $\pm$4.49} & \textbf{70.95} {\tiny $\pm$1.48} & \textbf{85.13} {\tiny $\pm$0.86} & \textbf{74.62} {\tiny $\pm$5.41} & \textbf{87.44} {\tiny $\pm$5.13} & \textbf{77.14} {\tiny $\pm$3.72} & \textbf{91.26} {\tiny $\pm$2.27} \\
\hline
\end{tabular}%
}
\end{table*}

\subsection{Ablation Study}
\label{subsec:ablation}

Table~\ref{tab:main_ablation} evaluates the main components by removing the WSI graph, SASGs, or SG-SSM. Each removal reduces performance across the evaluated tasks.
The improvement introduced by SASGs can be attributed to their ability to 
represent localized histopathological structures with irregular shapes and 
spatial coherence. Unlike regular grid-based patches, SASGs approximate 
semantic units while preserving their integrity and connectivity, allowing 
the model to focus on regional patterns rather than fragmented cues. The 
effectiveness of SG-SSM can be attributed to its ability to model long-range 
contextual relationships among SASGs, thereby facilitating the characterization 
of distinctive combinations of histological patterns for subtyping. Together, 
the two components support a comprehensive analysis of histological patterns 
and address the two coupled challenges discussed in 
Introduction. In addition, the performance degradation 
observed after removing the WSI graph suggests that complete slide coverage 
provides an important basis for modeling spatially distributed semantic units.

\begin{table}[!h]
\caption{
Ablation Study of Main Components. Combing Three Components Yields Improvements beyond Incremental Benefits of Each Component Alone.}
\label{tab:main_ablation}
\centering
\setlength{\tabcolsep}{2pt}
\renewcommand{\arraystretch}{1.05}
\begin{tabular}{llccc|ccc}
\hline
Dataset & Variants & G & SG & G$\times$M & F1 & AUC & ACC \\
\hline
\multirow{4}{*}{ESCA} 
& -w/o Graph 
&  & $\checkmark$ & $\checkmark$ 
& 84.25 {\tiny $\pm$6.77} 
& 92.65 {\tiny $\pm$3.01} 
& 82.18 {\tiny $\pm$7.77} \\

& -w/o SASG 
& $\checkmark$ &  & $\checkmark$ 
& 82.27 {\tiny $\pm$4.33} 
& 89.22 {\tiny $\pm$4.37} 
& 78.97 {\tiny $\pm$4.80} \\

& -w/o SG-SSM 
& $\checkmark$ & $\checkmark$ &  
& 82.09 {\tiny $\pm$4.91} 
& 85.73 {\tiny $\pm$2.78} 
& 78.27 {\tiny $\pm$5.88} \\

& \textbf{SASG-SSM} 
& $\checkmark$ & $\checkmark$ & $\checkmark$ 
& \textbf{92.17} {\tiny $\pm$4.16} 
& \textbf{95.71} {\tiny $\pm$3.12} 
& \textbf{91.09} {\tiny $\pm$4.61} \\
\hline

\multirow{4}{*}{BRCA} 
& -w/o Graph 
&  & $\checkmark$ & $\checkmark$ 
& 24.99 {\tiny $\pm$21.73} 
& 76.56 {\tiny $\pm$7.12} 
& 80.07 {\tiny $\pm$1.86} \\

& -w/o SASG 
& $\checkmark$ &  & $\checkmark$ 
& 26.82 {\tiny $\pm$16.79} 
& 75.46 {\tiny $\pm$5.17} 
& 78.83 {\tiny $\pm$3.01} \\

& -w/o SG-SSM 
& $\checkmark$ & $\checkmark$ &  
& 33.12 {\tiny $\pm$25.33} 
& 75.11 {\tiny $\pm$7.66} 
& 80.03 {\tiny $\pm$3.96} \\

& \textbf{SASG-SSM} 
& $\checkmark$ & $\checkmark$ & $\checkmark$ 
& \textbf{38.82} {\tiny $\pm$16.59} 
& \textbf{77.53} {\tiny $\pm$5.08} 
& \textbf{80.14} {\tiny $\pm$1.75} \\
\hline

\multirow{4}{*}{NSCLC} 
& -w/o Graph 
&  & $\checkmark$ & $\checkmark$ 
& 71.74 {\tiny $\pm$4.28} 
& 80.42 {\tiny $\pm$5.31} 
& 73.01 {\tiny $\pm$3.95} \\

& -w/o SASG 
& $\checkmark$ &  & $\checkmark$ 
& 70.33 {\tiny $\pm$2.68} 
& 77.02 {\tiny $\pm$3.67} 
& 68.73 {\tiny $\pm$3.72} \\

& -w/o SG-SSM 
& $\checkmark$ & $\checkmark$ &  
& 72.66 {\tiny $\pm$4.81} 
& 79.45 {\tiny $\pm$6.15} 
& 73.12 {\tiny $\pm$5.57} \\

& \textbf{SASG-SSM} 
& $\checkmark$ & $\checkmark$ & $\checkmark$ 
& \textbf{75.01} {\tiny $\pm$3.98} 
& \textbf{83.83} {\tiny $\pm$3.59} 
& \textbf{76.54} {\tiny $\pm$2.95} \\
\hline

\multirow{4}{*}{RCC} 
& -w/o Graph 
&  & $\checkmark$ & $\checkmark$ 
& 80.40 {\tiny $\pm$6.89} 
& 92.00 {\tiny $\pm$4.62} 
& 80.46 {\tiny $\pm$6.28} \\

& -w/o SASG 
& $\checkmark$ &  & $\checkmark$ 
& 78.76 {\tiny $\pm$4.45} 
& 91.85 {\tiny $\pm$2.01} 
& 78.86 {\tiny $\pm$4.15} \\

& -w/o SG-SSM 
& $\checkmark$ & $\checkmark$ &  
& 78.96 {\tiny $\pm$8.07} 
& 92.78 {\tiny $\pm$3.01} 
& 80.74 {\tiny $\pm$6.46} \\

& \textbf{SASG-SSM} 
& $\checkmark$ & $\checkmark$ & $\checkmark$ 
& \textbf{81.37} {\tiny $\pm$3.52} 
& \textbf{93.85} {\tiny $\pm$1.77} 
& \textbf{81.17} {\tiny $\pm$3.33} \\
\hline
\end{tabular}%
\end{table}

\subsection{Further Analysis}
\label{subsec:further_analysis_flag}

\subsubsection{Analysis of Semantic-Aware Subgraphs (SASGs)}
\label{subsubsec:sasg_analysis}

\mbox{}\par

\begin{figure}[!h]
\centering
\includegraphics[width=0.8\columnwidth]{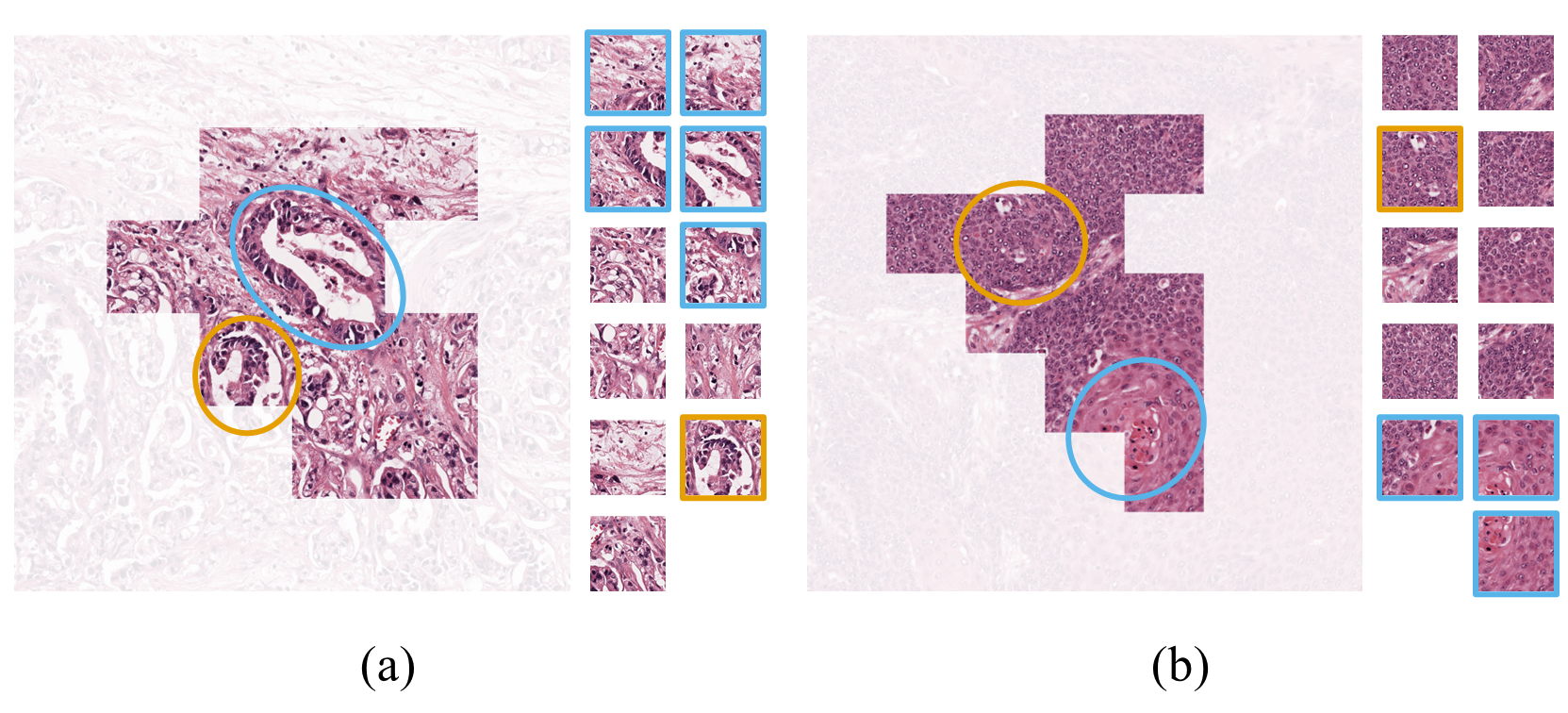}
\caption{Comparison between Semantic-Aware Subgraphs (SASGs) and patches in 
representing semantic units. SASGs approximate holistic histological structures 
that reflect local histological patterns, whereas individual patches provide 
fragmented and incomplete representations.}
\label{fig:sasg_visualization}
\end{figure}

\normalcolor
\textit{SASG Visualization:}
To examine the ability of SASGs to represent semantic units, we compare representative high-attention SASGs with individual patches. One EAD and one ESCC slide from TCGA-ESCA are shown in Fig.~\ref{fig:sasg_visualization}.

In Fig.~\ref{fig:sasg_visualization}(a), the EAD SASG preserves irregular glandular structures (yellow and blue circles) and their broader organization, whereas the corresponding structures are fragmented across isolated patches. 
In Fig.~\ref{fig:sasg_visualization}(b), the ESCC SASG captures solid and nested tumor architecture (yellow circle) together with concentric keratinization suggestive of keratin pearl formation~\cite{who_classification_of_tumours_editorial_board_digestive_2019} (blue circle). 
Individual patches provide only local fragments of these patterns. 
These examples indicate that connected multi-patch SASGs provide a more complete representation of local tumor architecture.

\begin{table}[!t]
\caption{
Comparison of Different Priors and Sampling Strategies for SASG Construction. SAM2-derived Prior and Adaptive Random Walk Sampling are Proven to be an Effective Implementation.}
\label{tab:sasg_construction}
\centering
\setlength{\tabcolsep}{2pt}
\renewcommand{\arraystretch}{1.15}

\begin{tabular}{lll|ccc}
\hline
Dataset & Prior & Sampling & F1 & AUC & ACC \\
\hline

\multirow{5}{*}{ESCA}
& No
& RW(Fixed)
& 89.08 {\tiny $\pm$4.02}
& 94.34 {\tiny $\pm$3.43}
& 87.28 {\tiny $\pm$4.05} \\
\cline{2-6}

& SLIC
& RW(Adaptive)
& 90.33 {\tiny $\pm$3.88}
& 94.39 {\tiny $\pm$2.49}
& 88.57 {\tiny $\pm$4.64} \\
\cline{2-6}

& \multirow{3}{*}{SAM2}
& \textbf{RW(Adaptive)}
& \textbf{92.17} {\tiny $\pm$4.16}
& \textbf{95.71} {\tiny $\pm$3.12}
& \textbf{91.09} {\tiny $\pm$4.61} \\
\cline{3-6}

&
& RW(Fixed)
& 90.10 {\tiny $\pm$2.47}
& 95.07 {\tiny $\pm$1.63}
& 88.51 {\tiny $\pm$3.04} \\
\cline{3-6}

&
& KN(Adaptive)
& 88.30 {\tiny $\pm$6.97}
& 93.94 {\tiny $\pm$3.43}
& 87.26 {\tiny $\pm$6.65} \\
\hline

\multirow{5}{*}{BRCA}
& No
& RW(Fixed)
& 26.66 {\tiny $\pm$23.22}
& 75.70 {\tiny $\pm$6.18}
& \textbf{81.06} {\tiny $\pm$0.88} \\
\cline{2-6}

& SLIC
& RW(Adaptive)
& 29.62 {\tiny $\pm$19.41}
& 72.48 {\tiny $\pm$8.96}
& 77.65 {\tiny $\pm$5.46} \\
\cline{2-6}

& \multirow{3}{*}{SAM2}
& \textbf{RW(Adaptive)}
& \textbf{38.82} {\tiny $\pm$16.59}
& \textbf{77.53} {\tiny $\pm$5.08}
& 80.14 {\tiny $\pm$1.75} \\
\cline{3-6}

&
& RW(Fixed)
& 30.21 {\tiny $\pm$15.38}
& 76.65 {\tiny $\pm$4.15}
& 79.11 {\tiny $\pm$1.34} \\
\cline{3-6}

&
& KN(Adaptive)
& 36.45 {\tiny $\pm$15.53}
& 77.31 {\tiny $\pm$5.17}
& 78.99 {\tiny $\pm$2.45} \\
\hline

\multirow{5}{*}{NSCLC}
& No
& RW(Fixed)
& 72.12 {\tiny $\pm$4.04}
& 78.53 {\tiny $\pm$6.97}
& 70.49 {\tiny $\pm$7.30} \\
\cline{2-6}

& SLIC
& RW(Adaptive)
& 74.47 {\tiny $\pm$2.82}
& 79.07 {\tiny $\pm$6.44}
& 71.72 {\tiny $\pm$7.20} \\
\cline{2-6}

& \multirow{3}{*}{SAM2}
& \textbf{RW(Adaptive)}
& \textbf{75.01} {\tiny $\pm$3.98}
& \textbf{83.83} {\tiny $\pm$3.59}
& \textbf{76.54} {\tiny $\pm$2.95} \\
\cline{3-6}

&
& RW(Fixed)
& 72.09 {\tiny $\pm$3.01}
& 79.71 {\tiny $\pm$4.54}
& 72.92 {\tiny $\pm$3.67} \\
\cline{3-6}

&
& KN(Adaptive)
& 72.46 {\tiny $\pm$4.31}
& 78.85 {\tiny $\pm$6.28}
& 71.02 {\tiny $\pm$6.75} \\
\hline

\multirow{5}{*}{RCC}
& No
& RW(Fixed)
& 79.27 {\tiny $\pm$5.81}
& 92.36 {\tiny $\pm$1.49}
& 80.44 {\tiny $\pm$4.40} \\
\cline{2-6}

& SLIC
& RW(Adaptive)
& 79.36 {\tiny $\pm$4.07}
& 91.83 {\tiny $\pm$2.31}
& 79.69 {\tiny $\pm$3.59} \\
\cline{2-6}

& \multirow{3}{*}{SAM2}
& \textbf{RW(Adaptive)}
& \textbf{81.37} {\tiny $\pm$3.52}
& \textbf{93.85} {\tiny $\pm$1.77}
& 81.17 {\tiny $\pm$3.33} \\
\cline{3-6}

&
& RW(Fixed)
& 80.17 {\tiny $\pm$4.59}
& 92.52 {\tiny $\pm$2.18}
& 80.55 {\tiny $\pm$4.25} \\
\cline{3-6}

&
& KN(Adaptive)
& 81.11 {\tiny $\pm$3.75}
& 92.20 {\tiny $\pm$1.70}
& \textbf{81.39} {\tiny $\pm$3.23} \\
\hline

\end{tabular}%
\end{table}

\begin{figure}[!h]
\centering
\includegraphics[width=0.8\columnwidth]{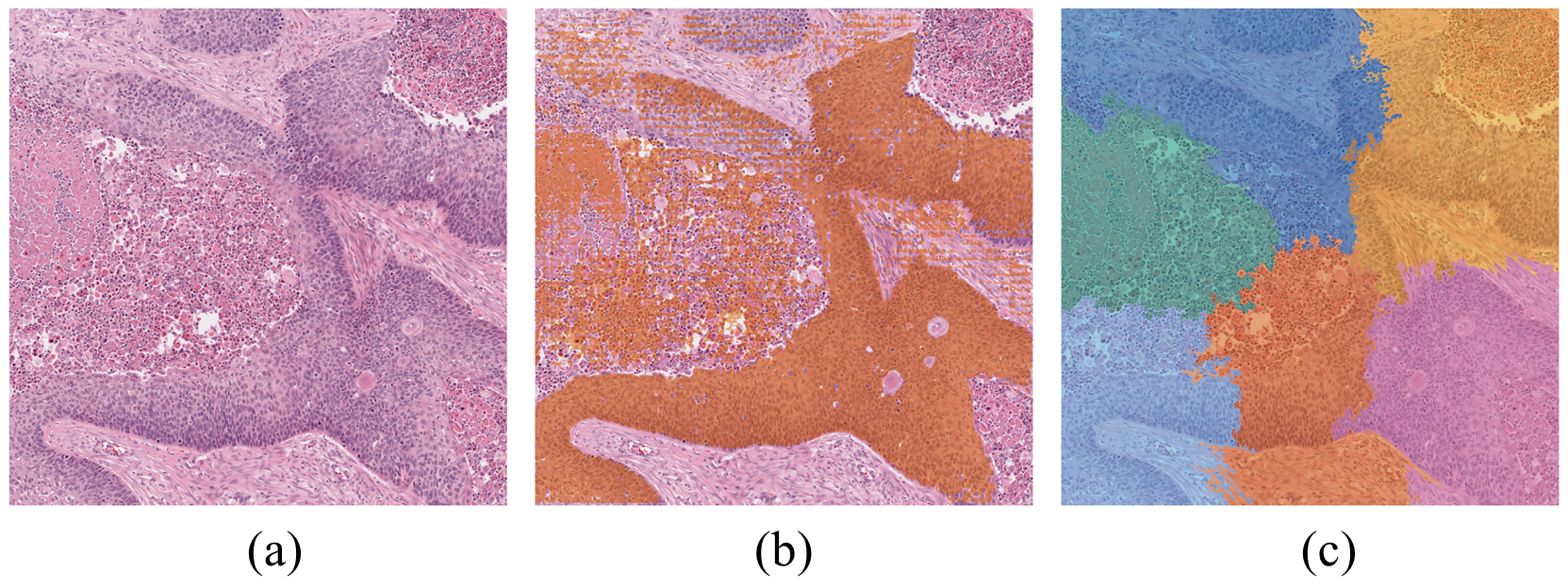}
\caption{Comparison of different visual-semantic priors for SASG construction. 
More informative priors facilitate the generation of coherent and discriminative 
SASGs. (a) Original image. (b) SAM2 segmentation mask. It seperates tumor nests from surrounding stromal and necrotic regions, providing more informative priors. (c) SLIC segmentation mask. It provides useful local boundaries but lacks comparable region-level visual information.
}
\label{fig:semantic_prior}
\end{figure}

\textit{Influence of Semantic Prior:}
We compare SAM2 with SLIC~\cite{achanta_slic_2012} and a ``No Prior'' setting while retaining the same subgraph-construction procedure. 
As shown in first 3 rows of each dataset in Table~\ref{tab:sasg_construction}, SAM2 achieves the strongest overall performance, indicating that more informative region priors improve SASG construction. 
Fig.~\ref{fig:semantic_prior}(b) shows that SAM2 better separates tumor nests from surrounding stromal and necrotic regions, enabling more coherent sampling. SLIC retains useful local boundaries but lacks comparable region-level visual information, whereas unconstrained sampling performs worst overall.

\textit{Influence of Adaptive Length:}
It is observed that replacing adaptive termination with a fixed walk length reduces performance (Table~\ref{tab:sasg_construction}). 
The ambiguity condition limits propagation near uncertain mask boundaries, while the no-candidate condition prevents repeated sampling. 
Together, these criteria allow the semantic prior to influence subgraph extent instead of imposing a fixed geometric size.

\textit{Influence of Sampling Strategy:}
We further compare $k$-hop neighborhood expansion (KN) and random-walk expansion (RW). 
Both use the same prior (SAM2) and adaptive termination criteria, but KN includes all eligible neighboring nodes at each step whereas RW progressively samples one neighbor. 
As shown in Table~\ref{tab:sasg_construction}, their comparable performance suggests that the visual-semantic prior and adaptive termination are more influential than the specific expansion rule. 
The slight advantage of RW may arise from greater stochastic sampling diversity.

\textit{Effects of Preserving Internal Topology:}
We remove all SASG edges while retaining the same constituent patches, yielding an equivalent sub-bag without explicit spatial relationships. 
As shown in Table~\ref{tab:internal_topology}, the topology-preserving subgraph performs better across all datasets, supporting the value of retaining intra-unit spatial structure.

\begin{table}[!h]
\caption{
Effect of Preserving Internal Topology in SASG Construction. With the Same Constituent Patches, Topology-preserving Subgraphs Outperform Sub-bags.}
\label{tab:internal_topology}
\centering
\setlength{\tabcolsep}{2pt}
\renewcommand{\arraystretch}{1.05}
\begin{tabular}{ll|ccc}
\hline
Dataset & Method & F1 & AUC & ACC \\
\hline
\multirow{2}{*}{ESCA} 
& Sub-bag 
& 80.70 {\tiny $\pm$5.62} 
& 82.11 {\tiny $\pm$5.25} 
& 76.98 {\tiny $\pm$6.21} \\

& \textbf{Subgraph} 
& \textbf{92.17} {\tiny $\pm$4.16} 
& \textbf{95.71} {\tiny $\pm$3.12} 
& \textbf{91.09} {\tiny $\pm$4.61} \\
\hline

\multirow{2}{*}{BRCA} 
& Sub-bag 
& 25.20 {\tiny $\pm$21.75} 
& 75.24 {\tiny $\pm$5.68} 
& 79.66 {\tiny $\pm$1.97} \\

& \textbf{Subgraph} 
& \textbf{38.82} {\tiny $\pm$16.59} 
& \textbf{77.53} {\tiny $\pm$5.08} 
& \textbf{80.14} {\tiny $\pm$1.75} \\
\hline

\multirow{2}{*}{NSCLC} 
& Sub-bag 
& 70.67 {\tiny $\pm$3.90} 
& 78.78 {\tiny $\pm$6.34} 
& 69.40 {\tiny $\pm$6.39} \\

& \textbf{Subgraph} 
& \textbf{75.01} {\tiny $\pm$3.98} 
& \textbf{83.83} {\tiny $\pm$3.59} 
& \textbf{76.54} {\tiny $\pm$2.95} \\
\hline

\multirow{2}{*}{RCC} 
& Sub-bag 
& 79.51 {\tiny $\pm$4.55} 
& 91.78 {\tiny $\pm$2.17} 
& 79.92 {\tiny $\pm$4.04} \\

& \textbf{Subgraph} 
& \textbf{81.37} {\tiny $\pm$3.52} 
& \textbf{93.85} {\tiny $\pm$1.77} 
& \textbf{81.17} {\tiny $\pm$3.33} \\
\hline
\end{tabular}%
\end{table}

\subsubsection{Analysis of Subgraph State Space Module (SG-SSM)}
\label{subsubsec:sgssm_analysis}

\mbox{}\par

\textit{Contextual Relationships Visualization:}
We reduce SG-SSM input and output representations into two-dimensions by UMAP~\cite{mcinnes_umap_2020} and identify dominant feature clusters by HDBSCAN~\cite{campello_density-based_2013} for visualization.

\begin{figure}[!h]
\centering
\includegraphics[width=0.8\columnwidth]{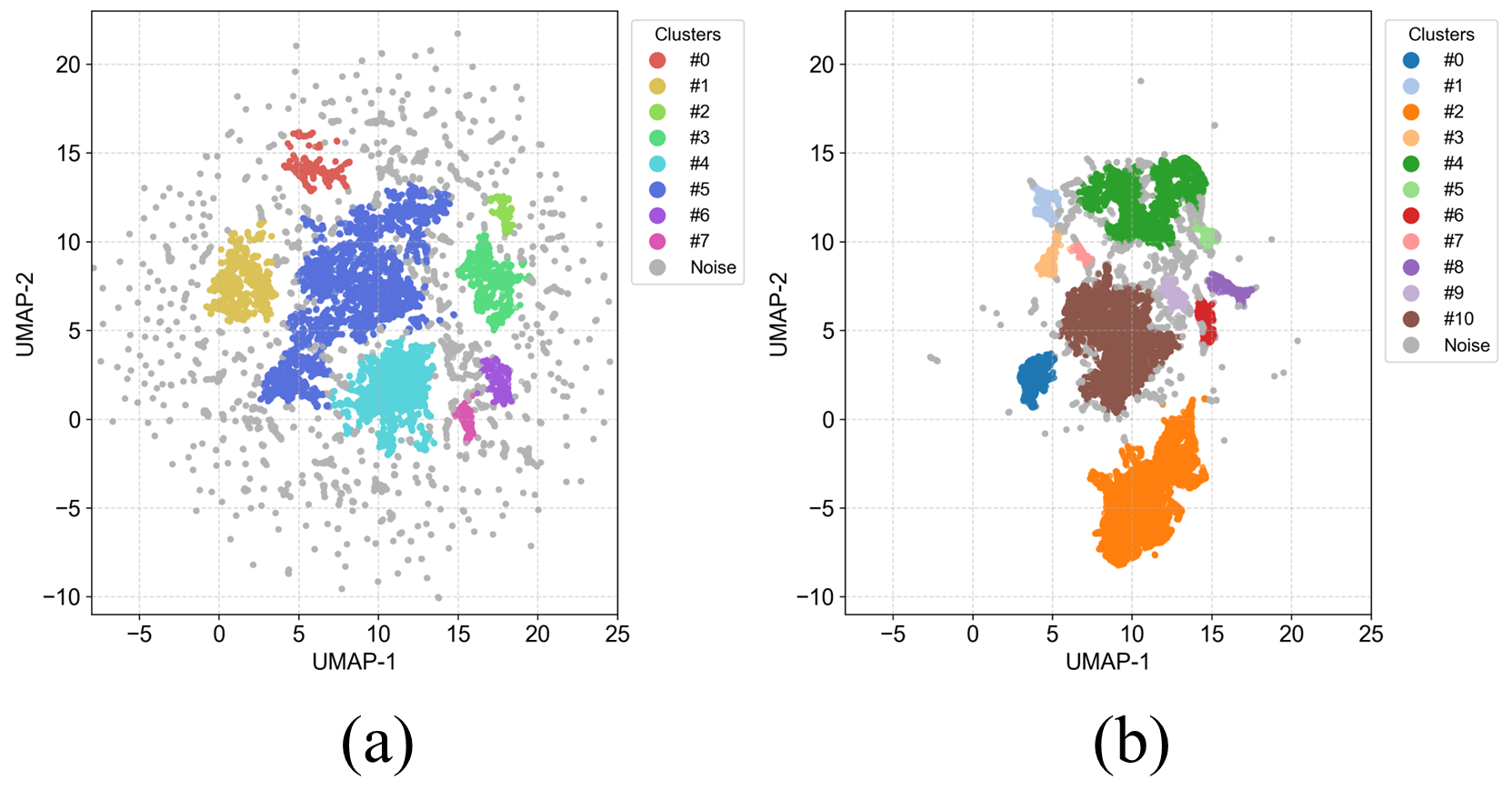}
\caption{UMAP visualization and HDBSCAN clustering of representations before 
and after SG-SSM. (a) Input representations of SG-SSM. The representations 
exhibit a relatively dispersed distribution, with a high proportion of samples 
identified as noise. (b) Output representations of SG-SSM. The refined 
representations form more compact and better-separated clusters, reflecting a 
more structured representation space.}
\label{fig:umap}
\end{figure}

\begin{figure}[!h]
\centering
\includegraphics[width=0.8\columnwidth]{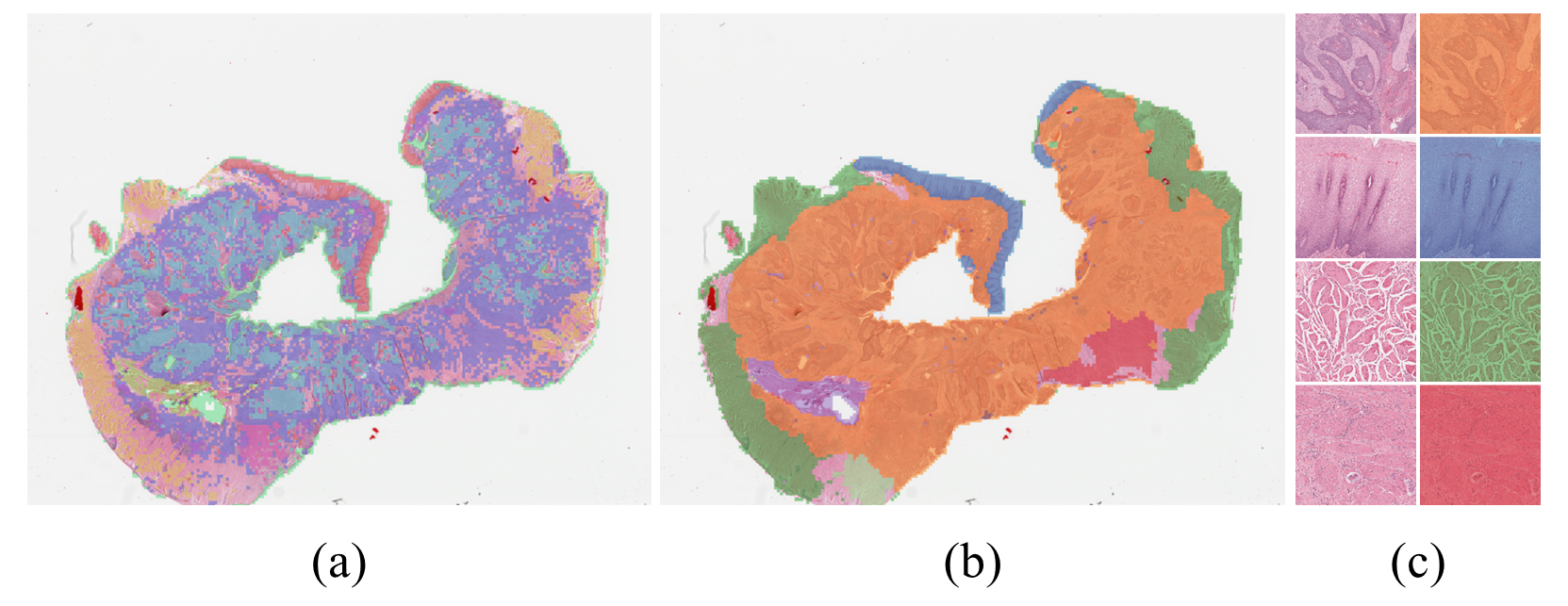}
\caption{Spatial projection of HDBSCAN cluster assignments onto the original 
WSI. (a) Spatial cluster map derived from the input representations of SG-SSM. 
Cluster assignments are relatively fragmented and spatially interwoven. 
(b) Spatial cluster map derived from the output representations of SG-SSM. 
The resulting clusters form more spatially coherent regions and exhibit 
distinct histomorphological patterns. (c) Representative regions cropped from 
(b), together with the corresponding original histological images. The orange 
cluster covers a tumor-infiltrated region containing diffusely distributed 
tumor nests. The blue cluster corresponds to an area with relatively preserved 
epithelial structures. The green and red clusters correspond predominantly to 
skeletal and smooth muscle tissues, respectively. The spatial arrangement of 
these tissue compartments may reflect an invasive pattern of esophageal 
squamous cell carcinoma.}
\label{fig:spatial_cluster}
\end{figure}

Fig.~\ref{fig:umap} displays the clustering map. Compared with the input, SG-SSM outputs form more compact and better-separated clusters with fewer noise samples in the embedding space. Fig.~\ref{fig:spatial_cluster} illustrates the spatial projection of cluster assignments back to the original WSI.
Output projections are also less fragmented and form more coherent regions while preserving boundaries between morphologically distinct tissue compartments.
Several clusters correspond to tumor-infiltrated, epithelial, and muscular regions. 
These observations provide qualitative evidence that SG-SSM integrates contextual information across related subgraphs and promotes a more structured representation space.

\begin{table}[!h]
\caption{
Ablation of GNN and SSM Encoders and Comparison of Their Incorporating Styles in SG-SSM.}
\label{tab:sgssm_ablation}
\centering
\setlength{\tabcolsep}{2pt}
\renewcommand{\arraystretch}{1.05}

\begin{tabular}{ll|ccc}
\hline
Dataset & Variant & F1 & AUC & ACC \\
\hline

\multirow{4}{*}{ESCA}
& -w/o GNN
& 81.65 {\tiny $\pm$4.74}
& 84.97 {\tiny $\pm$5.41}
& 78.29 {\tiny $\pm$5.84} \\

& -w/o SSM
& 85.24 {\tiny $\pm$6.19}
& 90.13 {\tiny $\pm$5.35}
& 82.84 {\tiny $\pm$5.90} \\

& Parallel
& 86.22 {\tiny $\pm$4.97}
& 87.86 {\tiny $\pm$4.39}
& 83.49 {\tiny $\pm$5.51} \\

& \textbf{Serial}
& \textbf{92.17} {\tiny $\pm$4.16}
& \textbf{95.71} {\tiny $\pm$3.12}
& \textbf{91.09} {\tiny $\pm$4.61} \\
\hline

\multirow{4}{*}{BRCA}
& -w/o GNN
& 34.63 {\tiny $\pm$20.77}
& 73.21 {\tiny $\pm$8.60}
& 79.27 {\tiny $\pm$4.23} \\

& -w/o SSM
& 31.27 {\tiny $\pm$21.65}
& 71.96 {\tiny $\pm$8.96}
& 79.69 {\tiny $\pm$3.10} \\

& Parallel
& 34.47 {\tiny $\pm$18.03}
& 74.90 {\tiny $\pm$4.51}
& 77.62 {\tiny $\pm$2.69} \\

& \textbf{Serial}
& \textbf{38.82} {\tiny $\pm$16.59}
& \textbf{77.53} {\tiny $\pm$5.08}
& \textbf{80.14} {\tiny $\pm$1.75} \\
\hline

\multirow{4}{*}{NSCLC}
& -w/o GNN
& 73.31 {\tiny $\pm$2.64}
& 82.33 {\tiny $\pm$6.83}
& 72.79 {\tiny $\pm$7.88} \\

& -w/o SSM
& 69.81 {\tiny $\pm$7.69}
& 78.37 {\tiny $\pm$6.17}
& 71.52 {\tiny $\pm$5.80} \\

& Parallel
& 71.45 {\tiny $\pm$2.67}
& 78.15 {\tiny $\pm$4.76}
& 70.50 {\tiny $\pm$4.84} \\

& \textbf{Serial}
& \textbf{75.01} {\tiny $\pm$3.98}
& \textbf{83.83} {\tiny $\pm$3.59}
& \textbf{76.54} {\tiny $\pm$2.95} \\
\hline

\multirow{4}{*}{RCC}
& -w/o GNN
& 80.82 {\tiny $\pm$5.15}
& 93.80 {\tiny $\pm$2.04}
& 80.84 {\tiny $\pm$5.18} \\

& -w/o SSM
& 78.08 {\tiny $\pm$9.33}
& 90.63 {\tiny $\pm$3.65}
& 79.41 {\tiny $\pm$7.46} \\

& Parallel
& 76.91 {\tiny $\pm$8.67}
& 91.40 {\tiny $\pm$4.01}
& 77.53 {\tiny $\pm$7.09} \\

& \textbf{Serial}
& \textbf{81.37} {\tiny $\pm$3.52}
& \textbf{93.85} {\tiny $\pm$1.77}
& \textbf{81.17} {\tiny $\pm$3.33} \\
\hline

\end{tabular}
\end{table}

\textit{Effects of GNN and SSM Encoders:}
Removing either the GNN or SSM encoder reduces performance in Table~\ref{tab:sgssm_ablation} (-w/o GNN \& -w/o SSM), supporting their complementary roles in modeling local topology and long-range context.

\textit{Influence of Incorporation Styles:}
Table~\ref{tab:sgssm_ablation} (Parallel \& Serial) also demonstrates the superiority of serial style when incorporating two encoders, suggesting that simply adding the two types of features is less effective than progressively updating the representations across graph and sequence spaces. 
This indicates that topology encoding followed by contextualization provides a more effective local-to-global flow.

\textit{Influence of Scan Order in the SSM Encoder:}
\label{sec:scan_order}
Since Mamba-based models can be sensitive to input sequence order, we further  
evaluate the influence of scan order in the SSM encoder of SG-SSM~\cite{yang_mambamil_2024}.
We evaluate two spatial scan directions and the ordering between multi-patch and zero-step subgraphs, with results shown in Table~\ref{tab:scan_order}.
Placing multi-patch subgraphs first consistently performs better, possibly because their richer regional information establishes a more informative contextual state before finer-grained zero-step tokens. 
Changing spatial direction has a smaller effect, consistent with the absence of a canonical WSI orientation.

\begin{table}[!h]
\caption{
Comparison of Scan Orders in SG-SSM. Multi-patch Subgraphs are Generally Preferred before Zero-step Subgraphs, whereas Performance is Relatively Insensitive to Spatial Directions.}
\label{tab:scan_order}
\centering
\setlength{\tabcolsep}{2pt}
\renewcommand{\arraystretch}{1.05}
\begin{tabular}{lll|ccc}
\hline
Dataset & Direction & Before & F1 & AUC & ACC \\
\hline
\multirow{4}{*}{ESCA} 
& \textbf{LT-BR} & \textbf{Multi-patch} 
& \textbf{92.17} {\tiny $\pm$4.16} 
& \textbf{95.71} {\tiny $\pm$3.12} 
& \textbf{91.09} {\tiny $\pm$4.61} \\

& BR-LT & Multi-patch 
& 89.40 {\tiny $\pm$4.07} 
& 95.05 {\tiny $\pm$2.32} 
& 87.92 {\tiny $\pm$4.25} \\

& LT-BR & Zero-step 
& 75.44 {\tiny $\pm$2.89} 
& 77.53 {\tiny $\pm$4.94} 
& 68.81 {\tiny $\pm$4.11} \\

& BR-LT & Zero-step 
& 70.54 {\tiny $\pm$9.63} 
& 64.76 {\tiny $\pm$11.69} 
& 65.08 {\tiny $\pm$10.07} \\
\hline

\multirow{4}{*}{BRCA} 
& \textbf{LT-BR} & \textbf{Multi-patch} 
& \textbf{38.82} {\tiny $\pm$16.59} 
& \textbf{77.53} {\tiny $\pm$5.08} 
& \textbf{80.14} {\tiny $\pm$1.75} \\

& BR-LT & Multi-patch 
& 36.74 {\tiny $\pm$16.76} 
& 77.35 {\tiny $\pm$5.09} 
& 79.58 {\tiny $\pm$2.25} \\

& LT-BR & Zero-step 
& 7.75 {\tiny $\pm$7.47} 
& 51.60 {\tiny $\pm$3.18} 
& 75.66 {\tiny $\pm$7.90} \\

& BR-LT & Zero-step 
& 9.13 {\tiny $\pm$5.80} 
& 53.16 {\tiny $\pm$1.78} 
& 75.36 {\tiny $\pm$1.18} \\
\hline

\multirow{4}{*}{NSCLC} 
& \textbf{LT-BR} & \textbf{Multi-patch} 
& \textbf{75.01} {\tiny $\pm$3.98} 
& \textbf{83.83} {\tiny $\pm$3.59} 
& \textbf{76.54} {\tiny $\pm$2.95} \\

& BR-LT & Multi-patch 
& 72.17 {\tiny $\pm$3.96} 
& 78.35 {\tiny $\pm$6.13} 
& 70.52 {\tiny $\pm$6.46} \\

& LT-BR & Zero-step 
& 41.74 {\tiny $\pm$7.48} 
& 52.27 {\tiny $\pm$4.44} 
& 52.54 {\tiny $\pm$3.37} \\

& BR-LT & Zero-step 
& 51.75 {\tiny $\pm$4.88} 
& 50.77 {\tiny $\pm$2.63} 
& 52.15 {\tiny $\pm$1.79} \\
\hline

\multirow{4}{*}{RCC} 
& \textbf{LT-BR} & \textbf{Multi-patch} 
& \textbf{81.37} {\tiny $\pm$3.52} 
& \textbf{93.85} {\tiny $\pm$1.77} 
& \textbf{81.17} {\tiny $\pm$3.33} \\

& BR-LT & Multi-patch 
& 78.78 {\tiny $\pm$5.48} 
& 92.31 {\tiny $\pm$1.53} 
& 80.05 {\tiny $\pm$4.15} \\

& LT-BR & Zero-step 
& 54.32 {\tiny $\pm$6.30} 
& 65.46 {\tiny $\pm$6.63} 
& 58.54 {\tiny $\pm$4.06} \\

& BR-LT & Zero-step 
& 49.47 {\tiny $\pm$4.43} 
& 64.25 {\tiny $\pm$3.92} 
& 54.20 {\tiny $\pm$2.48} \\
\hline
\end{tabular}%
\end{table}

\section{Conclusion}
In this work, we proposed Semantic-Aware Subgraph State Space Model for histopathological WSI subtyping. 
Semantic-Aware Subgraphs approximate irregularly shaped candidate semantic units spanning across multi-patches, preserving spatial organization with internal topology rather than unordered collections of patches.
Subgraph State Space Module employs GNN-based topology encoding before Mamba-based long-range contextualization, establishing a local-to-global information flow across the WSI. 

Experiments on four TCGA cohorts demonstrate strong overall performance, particularly under small-cohort and few-shot settings, while ablation and qualitative analyses support the contributions of prior-guided subgraph construction, topology preservation, and local-to-global contextual modeling.

The current implementation uses SAM2-derived visual-semantic priors and mask-constrained adaptive random walks for SASG construction.
Future work may explore pathology-specific priors and learnable subgraph construction, evaluate external multi-center cohorts, and extend SASG-SSM to more complex clinical tasks such as histological grading and staging.



\end{document}